\documentclass[journal]{IEEEtran}

\usepackage{graphicx}
\usepackage{bm}
\usepackage[caption=false,font=footnotesize]{subfig}
\usepackage{multirow}
\usepackage{flushend}
\usepackage{cite}
\usepackage{stfloats}

\newcommand{\mystep}[1]{{\vspace{2mm}\noindent\textbf{#1}}}

\begin{document}

\title{Touch Sensors with Overlapping Signals: Concept Investigation on Planar Sensors with Resistive or Optical Transduction}

\author{Pedro~Piacenza$^{1}$, Emily Hannigan$^{1}$, Clayton~Baumgart$^{1}$, Yuchen~Xiao$^{1}$, Steve~Park$^{3}$, Keith~Behrman$^{2}$, Weipeng~Dang$^{2}$, Jeremy~Espinal$^{4}$, Ikram~Hussain$^{4}$, Ioannis~Kymissis$^{2}$ and Matei~Ciocarlie$^{1}$% <-this % stops a space

\thanks{$^{1}$Department of Mechanical Engineering, Columbia University, New York, NY 10027, USA.}%
\thanks{\hspace{-3mm}{\tt\small \{pp2511,ejh2192,cgb2132,yx2281, matei.ciocarlie\}@columbia.edu}}%
\thanks{$^{2}$Department of Electrical Engineering, Columbia University, New York, NY 10027, USA.}%
\thanks{\hspace{-3mm}{\tt\small \{kb2902,wd2265,johnkym@ee\}@columbia.edu}}%
\thanks{$^{3}$Department of Materials Science and Engineering, Korea Advanced Institute of Science and Technology, Daejeon, Korea}%
\thanks{\hspace{-3mm}{\tt\small stevepark@kaist.ac.kr}}%
\thanks{$^{4}$Columbia Engineering ENG Summer Research Program.}}

\maketitle

\begin{abstract}
Traditional methods for achieving high localization accuracy on tactile sensors usually involve a matrix of miniaturized individual sensors distributed on the area of interest. This approach usually comes at a price of increased complexity in fabrication and circuitry, and can be hard to adapt to non-planar geometries. We propose a method where sensing terminals are embedded in a volume of soft material. Mechanical strain in this material results in a measurable signal between any two given terminals. By having multiple terminals and pairing them against each other in all possible combinations, we obtain a rich signal set using few wires. We mine this data to learn the mapping between the signals we extract and the contact parameters of interest. Our approach is general enough that it can be applied with different transduction methods, and achieves high accuracy in identifying indentation location and depth. Moreover, this method lends itself to simple fabrication techniques and makes no assumption about the underlying geometry, potentially simplifying future integration in robot hands.

\end{abstract}


\section{Introduction}

Tactile sensing modalities for robotic manipulation have made great
strides over the past years. Numerous transduction methods have been
explored: piezoresistance, piezocapacitance, piezoelectricity, optics,
ultrasonics, etc. Still, these advances in sensing modalities are only
slowly translating to improved abilities for complete robotic tactile
systems. A possible reason is that the gap between an individual taxel
and a fully fleshed out tactile system often proves difficult to
cross. As Dahiya et al.\cite{dahiya2010} conclude in an extensive survey, ``[w]hile new
tactile sensing arrays are designed to be flexible, conformable, and
stretchable, very few mention system constraints like [...]  embedded
electronics, distributed computing, networking, wiring, power
consumption, robustness, manufacturability, and
maintainability.''

\begin{figure}[t]
\centering
\includegraphics[clip, trim=4.8cm 1.2cm 4.8cm 1.2cm,
  width=0.75\linewidth]{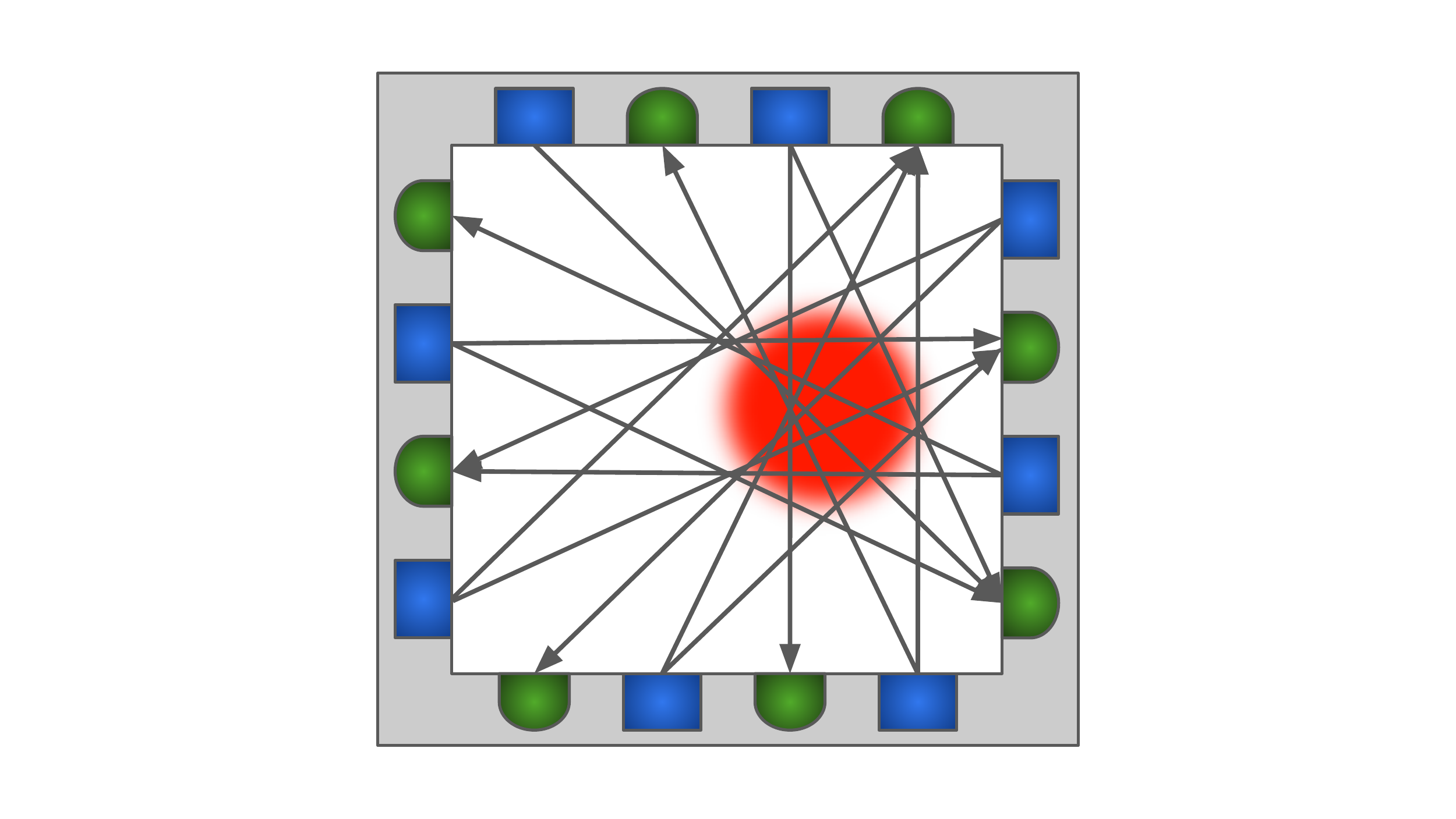}
\caption{Spatially Overlapping Signals. Arrows represent the spatial
  receptive field of a given pair of sensing terminals. A single
  contact affects signals extracted from multiple pairs
  whose receptive fields overlap on the contact location.}
\label{fig:overlapping_sig}
\end{figure}

In this paper we focus on the problem of developing a tactile sensing
system that addresses some of the challenges mentioned
above. Specifically we intend to reduce the electronic complexity in
terms of both wiring and manufacturing, while keeping the system cost
effective and achieving good coverage and accuracy for contact
localization and indentation depth.

Traditionally, high accuracy contact localization can be achieved
using dense arrays of individual taxels. This approach comes at a cost
of increased manufacturing complexity as individual taxels must be
isolated from each other. High resolution arrays normally lead to a
high number of sensing elements, which results in increased wiring and
can pose a challenge in terms of sampling all taxels quickly
enough. Moreover, deploying this method on non-planar surfaces
increases manufacturing complexity even further. Electric impedance
tomography (EIT) can construct a map of the sensor deformation with
relatively few wires placed on its periphery, but relies on the
existance of an analytical model of the underlying transduction method
as well as accurate sensor placement.

Our approach also starts by embedding and distributing multiple
sensing terminals in a volume of soft material. We assume a
transduction method such that strain in the material will measurably
affect a signal between two such terminals. For example, when using
piezoresistance, a strain in the material will change the resistance
between two terminals. In general, we expect that the change in the
measured signals is correlated with both the location and the
magnitude of the applied strain. However, we do not require an
analytical model of this correlation. This framework is general enough
that there are multiple transduction methods that can be used.

Now consider multiple such terminals embedded in the soft volume:
depending on their spatial distribution, a single contact event will
affect signals between multiple pairs (see
Fig.~\ref{fig:overlapping_sig}). The magnitude change of all these
signals contains information about the contact. A key aspect to
extract as much information as possible from this configuration is to
measure all possible terminal pairs: a single terminal can be measured
against many others. This rich data set can then be mined in purely
data-driven fashion to learn the mapping between these signals and the
contact parameters of interest. Because each pair of sensing terminals
has its own spatial receptive field within the active sensing area and
these receptive fields overlap, we call this general method
\textit{spatially overlapping signals}. If the mapping is learned
based strictly on collected data, we do not require exact knowledge of
terminal locations within the sensor; we simply require that
overlapping pairs provide coverage of the intended sensing area (we
quantify the impact of coverage on performance later in the paper).

Our method can be summarized into three key ideas, which are also the
main contributions of this paper:
\begin{itemize}
\item \textit{Relatively few terminals can give rise to many terminal
  pairs}, and measuring signal change between all these pairs provides
  a very rich data set characterizing touch.
\item \textit{We can use data-driven methods to directly learn the
  mapping between this rich signal set and our variables of interest},
  such as the location and depth of an indentation on the surface of
  the sensor. There is no need for complex models of the material
  properties or material deformation.
\item \textit{This approach is general enough to apply to multiple
  underlying transduction mechanisms.} We demonstrate it here with two
  such methods, based on piezoresistance and optics.
\end{itemize}
As we will detail in this paper, our method achieves spatial accuracy
in the order of 1 mm and less than 1 mm for depth accuracy. This is done
using few sensing terminals, which in turn reduces the number of wires
needed. We make no assumptions about the underlying sensor geometry,
and do not require precise terminal positioning inside the sensor. As
a result, the sensors presented in this paper can be manufactured at a
very low cost, and do not require complex electronics for signal
processing or for the sensing terminals themselves. These
characteristics make this method a good candidate to be integrated
into robot hands exhibiting complex 3D geometry, which is our
directional goal.

Initial versions of the work presented here on the design and
performance of the resistive and optical sensors respectively have
appeared in conference
proceedings:~\cite{piacenza2016,piacenza2017}. In this study, we show
a significant new capability: discriminating between multiple indenter
tips, each with a different shape. More importantly, we show that our
data-driven method still allows us to localize touch and determine the
depth of indentation accurately for an indenter geometry that the
sensor has not been trained on. To achieve this, we introduce a second
version of our optical sensor, using surface-mounted components, which
increases robustness to drift and allows us to diversify terminal
placement. Finally, we also investigate how the number and
distribution of sensing terminals affects performance, and how our
approach scales to larger sensors.

\section{Related Work}

Numerous types of transduction principles have been explored during
the last two decades when building tactile sensors. We refer the
reader to a number of comprehensive
reviews:~\cite{dahiya2010,hammock2013,kappassov2015} for an overview
of these methods. Our goal however is not to explore a new sensing
modality; rather, we are looking to build on top of such methods, in
order to provide continuous sensing with high accuracy without
sacrificing manufacturability.

Regardless of the base transduction principle, attempts to increase
spatial resolution have often resulted in the arrangement of multiple
discrete sensors into a matrix to cover a given target surface. Kane
et al.~\cite{kane2000}, Takao et al.~\cite{takao2006} and Suzuki et
al.~\cite{suzuki1990} have reported sensor arrays that can develop
very high spatial resolution. However, a drawback of this approach is
the difficulty involved in manufacturing these arrays onto a flexible
substrate than can conform to complex surfaces. Shimojo et
al.~\cite{shimojo2004} and Kim et al.~\cite{kim2008} present possible
technologies to overcome this problem like organic FETs/thin film
transistors realized on elastomeric substrates and other related
techniques. Still, wiring and manufacturing complexity, along with
other system-level issues such as addressing and signal processing of
multiple sensor elements, remain important roadblocks on the way to
building complete sensing systems.

Electric impedance tomography (EIT) is used to estimate the internal
conductivity of an electrically conductive body by virtue of
measurements taken with electrodes placed on the boundary of said
body. While originally used for medical applications, EIT techniques
have been applied successfully for manufacturing stretchable,
sensitive skin for robotics that can be deployed on complex surfaces
(see \cite{nagakubo2007,kato2007,tawil2011_improved,lee2017}). A
comprehensive survey on the use of EIT for robotic skin can be found
in the work of Silvera et al.~\cite{silvera2015}.

EIT skins are one of the few methodologies to offer stretchable,
continuous tactile sensing with the ability to discriminate multiple
points of contact with force sensing.  Like EIT, we also combine
signals from multiple emitter-receiver pairs.  However, EIT methods
require an analytical model for the internal conductivity to construct
an image showing the areas where strain is applied. In contrast, our
approach is completely data-driven and does not require the knowledge
of a forward model of the sensor. This is what allows us to formulate
this method in a general, transducer-agnostic fashion, which in turn
can allow the use of transduction methods that are easy to
manufacture, but difficult to model analytically. We demonstrate the
feasibility and generality of this model-free approach here by using
both piezoresistance and light transport as the underlying
transduction methods.

An intrinsic advantage of EIT is that it can produce full contact maps
for multi-touch situations, an ability which we have not yet
investigated with our method.  Instead, using our model-free approach,
we demonstrate and quantify the ability to very accurately localize
single touch (with both a known or unknown indenter), to determine
indentation depth, and to discriminate between different indenter
shapes. For grasping applications, we believe that the combination of
these abilities can prove valuable; for example, when manipulating
locally convex objects using fingers with convex links, each link will
establish a single contact with the target.  In other applications,
such full body coverage, where detecting multitouch is a strict
requirement, EIT represents a promising alternative.

Our localization approach leverages the overlap of receptive fields in a similar fashion 
to methods such as super-resolution and electric impedance
tomography. van den Heever et al.~\cite{heever2009} used a similar algorithm to
super-resolution imaging, combining several measurements of a 5 by 5
force sensitive resistors array into an overall higher resolution
measurement. Lepora and Ward-Cherrier~\cite{lepora2015_superresolution} and Lepora et
al.~\cite{lepora2015_tactile} used a Bayesian
perception method to obtain a 35-fold improvement of localization
acuity (0.12mm) over a sensor resolution of 4mm. In general,
super-resolution techniques for tactile sensing leverage overlapping
receptive fields of neighboring taxels to perceive stimuli detail
finer than the sensor resolution. In this context the sensor
resolution relates to the spacing between taxels. While our approach
also leverages the overlap of receptive fields, in our case there is
no obvious resolution metric for the sensor, since our signals are not
the result of an individual taxel, but of a terminal pair whose
receptive field is determined by the pair location. 

Other sensors also use a small number of underlying transducers to
recover richer information about the contact. For example, work in the
ROBOSKIN project showed how to calibrate multiple piezocapacitive
transducers~\cite{cannata2010}, used them to recover a complete
contact profile~\cite{muscari2013} using an analytic model of
deformation, and finally used such information for manipulation
learning tasks~\cite{argall2011}. Our localization method is entirely
data driven and makes no assumptions about the underlying properties
of the medium, which could allow coverage of more complex geometric
surfaces. Hosoda et al.~\cite{hosoda2006} built an anthropomorphic 
soft finger with randomly distributed strain gauges and PVDF
(polyvinylidene fluoride) films to discriminate between five materials by pressing 
and rubbing them. This study shares a similar vision to ours because it relies on
embedded sensing terminals inside a soft volume, but the authors did not 
explore the possibility of localizing touch or measuring the contact force. 
A human operator chooses which pairs carry relevant information and then perform
an analytical study of the signal variance to achieve the material classification 
instead of a learning approach. 

The basic building block for one of the sensors presented here is an
elastomer with dispersed conductive fillers applied to achieve
piezoresistive characteristics. Chortos and Bao~\cite{chortos2014}, Dusek et al.~\cite{dusek2014} and
Kim et al.~\cite{kim2015} are just some examples in the literature of this
methodology, with carbon black and carbon nanotubes as the most
commonly employed fillers. Mannsfeld et al.~\cite{mannsfeld2010}, Park et al.~\cite{park2014} and
Wu et al.~\cite{wu2015} have shown that multi-layered designs or additional
microstructures can further improve performance. Park et al.~\cite{park2012} and Vogt et al.~\cite{vogt2013} reported that embedding microchannels of conductive
fluids inside an elastic volume can be an effective alternative to
making the entire volume conductive, especially if large strains are
desirable. Here however we opt for the simplicity of single volume
isotropic material which can be directly molded into the desired
shape. The use of conductive fillers dispersed on elastomers has
previously been used to develop tactile
sensors. Charalambides et al.~\cite{charalambides2015} used carbon infused PDMS to develop
a 3-axis MEMS tactile sensor, using capacitance to measure the
deformation of carefully designed pillars.

The other sensors presented in this paper use light transport through
an elastomer as a transduction method. The use of optics for tactile
sensing is not new, and has a long history of integration in robotic
fingers and hands. Early work by Begej~\cite{begej1988} demonstrated the use
of CCD sensors recording light patterns through a robotic tip affected
by deformation. More recently Lepora and Ward-Cherrier~\cite{lepora2015_superresolution} showed how to achieve
super-resolution and hyperacuity with CCD-based touch sensors
integrated into a fingertip. Schneider et al.~\cite{schneider2009} used color-coded 3D
geometry reconstruction to retrieve minute surface details with an
optics-based sensor. These studies share a common concept of a CCD
array imaging a deformed fingertip from the inside, requiring that the
array be positioned far enough from the surface in order to image the
entire touch area. In our optic sensor, the sensing terminals are
fully distributed, allowing for coverage of large areas and
potentially irregular geometry. Additionally, we take advantage of
multiple modes of light transport through an elastomer to increase the
sensitivity of the sensor. In recent work, Patel et al.~\cite{patel2017} took
advantage of reflection and refraction to build an IR touch sensor
that also functions as a proximity sensor. Their work however does not
provide means to also localize contact. Work by Polygerinos et al.~\cite{polygerinos2010}
uses the deformation of an optic fiber to create a force
transducer. This approach has the advantage that the sensing
electronics do not have to be located close to the contact area. A very
similar sensor exploiting the same physical principles was presented
by Levi et al.~\cite{levi2013} using an algorithm similar to tomographic
back-projection to reconstruct the surface deformation. Our method is
completely data driven and takes advantage of two different light
transport modes. While their work reports a superior sensitivity to
light contact, they do not present any accuracy data in terms of
contact localization.

We rely on data-driven methods to learn the behavior of our sensors;
along these lines, we note that machine learning for manipulation
based on tactile data is not new. Wong et al.~\cite{wong2014} learned to
discriminate between different types of geometric features based on
the signals provided by a previously developed multimodal touch sensor
by Wettels et al.~\cite{wettels2008}. Current work by Wan et al.~\cite{wan2016} relates tactile
signal variability and predictability to grasp stability using
recently developed MEMS-based sensors by Tenzer et al.~\cite{tenzer2014}. With
traditional tactile arrays, Dang and Allen\cite{allen2013} successfully used an SVM
classifier to distinguish stable from unstable grasps in the context
of robotic manipulation using a Barrett Hand. Silvera-Tawil \cite{tawil2011_interpretation}
managed to classify six different touch gestures on an experimental EIT-based skin.
In contrast, we apply data-driven methods to learn a model of the sensor itself 
and believe that developing the sensor simultaneously with the learning techniques that
make use of the data can bring us closer to achieving complete tactile
systems.

\section{Spatially Overlapping Signals}

\begin{figure*}[t]
\centering
\includegraphics[clip, trim=0cm 0cm 0cm 0cm,
  width=1.0\linewidth]{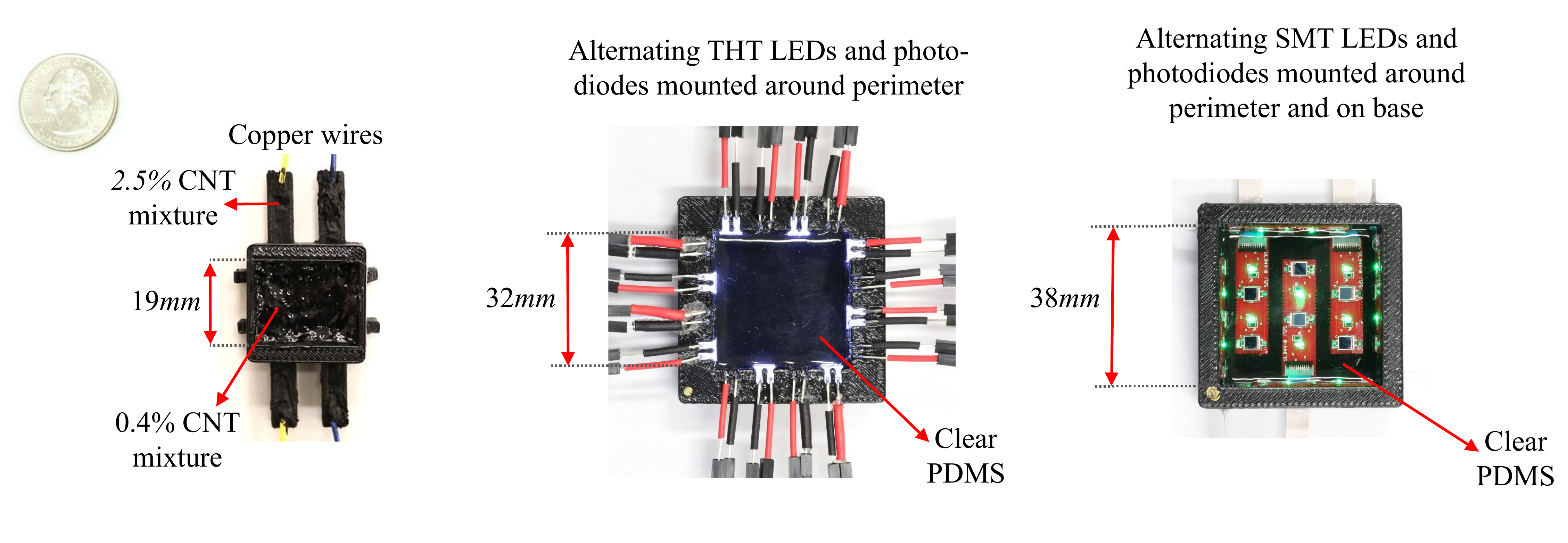}
\caption{Design of our sensors. \textbf{Left:} resistive sensor. The
  center is filled with piezoresistive PDMS/CNT mixture. Side channels
  are filled with a conductive mixture with higher CNT ratio in order
  to mechanically isolate wire contacts from
  indentations. \textbf{Middle:} optical sensor. Alternating
  through-hole technology (THT) LEDs and photodiodes are edge-mounted
  and the cavity is filled with a clear elastomer. \textbf{Right: }
  optical sensor with surface-mount technology (SMT) LEDs and
  photodiodes, mounted both along the perimeter and on the base.}
\label{fig:three_sensors}
\end{figure*}

Our approach begins with a continuous volume of soft material. We use
Polydimethylsiloxane (PDMS), a widely used silicon-based organic
polymer. It is optically clear and its stiffness can be adjusted by
changing the the ratio of curing agent to PDMS. In all sensors
presented here we used a ratio of 1:20 curing agent to PDMS by weight.

The spatially overlapping signals methodology is fundamentally based
on having multiple pairs of sensing terminals distributed or embedded
in the intended sensing area within a given volume of soft
material. Generally speaking, a single pair of sensing terminals has a
signal associated with it. The signal measured by this sensing pair
needs to be sensitive to applied mechanical strain on the underlying
material.

A given sensing pair has an associated receptive field or area of
influence. Applying mechanical strain within this area will produce a
measurable change in the signal for that particular sensing pair. We
embed and distribute multiple such sensing terminals within the soft
material, so that the intended active sensing area is covered with the
receptive fields of multiple sensing pairs. These receptive fields
will typically overlap each other, hence a single contact event will
have an effect on more than one sensing pair signal
(Fig.~\ref{fig:overlapping_sig}).

To maximize the data extracted we use an all-pairs approach, where all
possible sensing terminals pairs are used. This allows us to obtain a
very rich data set with few sensing terminals, resulting in a reduced
number of wires for the overall sensor. The number of sensing pairs
(and thus signals we harvest) is generally quadratic in the number of
terminals. This rich data set lends itself to the use of data-driven
techniques to directly learn the mapping between the signal set
extracted from the sensor and our variables of interest.

The spatially overlapping signals concept itself is transparent when
it comes to the sensing transduction method used. In this paper we
show this concept applied with two very distinct transduction methods:
resistivity and optics. It must be noted that how we construct these
sensing pairs varies depending on the transduction method. In our
resistive sensor (Fig.~\ref{fig:three_sensors}, Left), the sensing
terminal is simply an electrode and we can match a given terminal
with any other to measure resistance. However in the case of our
optical sensors (Fig.~\ref{fig:three_sensors}, Middle and Right), a
given sensing terminal has a defined role: it can either be an emitter
(an LED) or a receiver (a photodiode). An immediate consequence of
this fact is that the number of sensing pairs as you add a sensing
terminal grows slower for our optical sensors than for the resistive
sensor (though the number of pairs is still quadratic in the number of
terminals in both cases). Moreover, the resistive sensor only requires
one wire per sensing terminal, compared to the two wires required on
the optical sensors for each LED and photodiode.

One of the trade-offs in our method is that of data
processing. Building an analytical model of how this rich signal set
is affected by contact characteristics is a daunting task;
furthermore, any such model would depend on knowing the exact
locations of the terminals in the sensor, thus requiring very precise
manufacturing. Qualitatively, we expect that the intensity of
the collected signals is directly related to the location and
magnitude of the applied strain. 

\begin{figure}[t]
\centering
\includegraphics[clip, trim=0.3cm 0.5cm 0.3cm 0.5cm, width=0.65\linewidth]{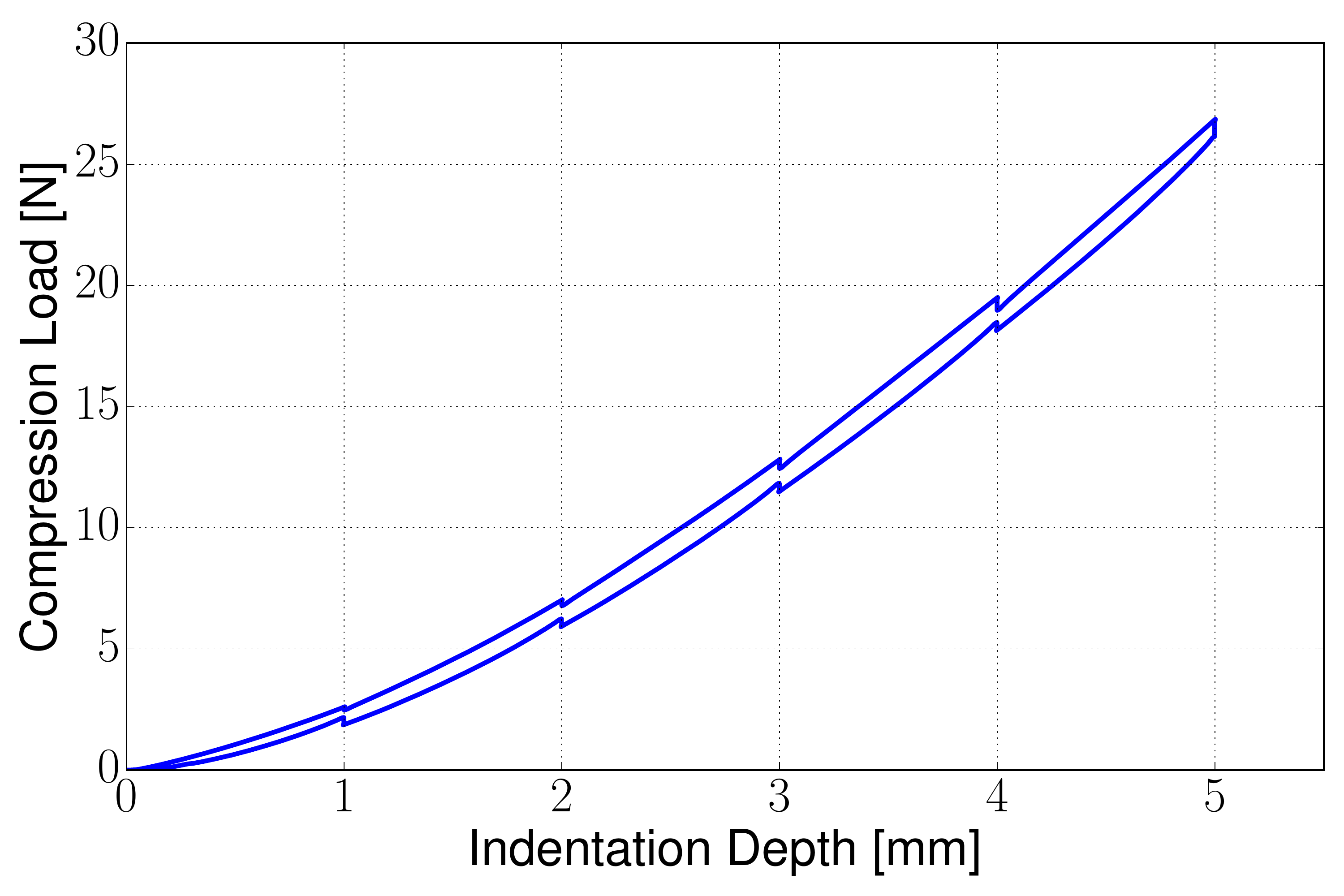}
\caption{Load vs indentation depth for a 1:20 ratio of curing agent to
  PDMS, measured by advancing or retracting the probe in 1mm steps
  with 10s pauses using a 6mm diameter hemispherical tip. }
\label{fig:Mapping}
\end{figure}

We thus use a purely data-driven method to directly learn the mapping
between the signal set extracted from the sensor and our variables of
interest. We train our method using a set of indentations of known
characteristics. In particular, in this paper we focus mainly on
learning a mapping for contact localization and indentation
depth. Throughout this study we use indentation depth as a proxy for
contact force; for conversion to force values, we provide the
stiffness curve of our constituent material in
Fig.~\ref{fig:Mapping}. Using an optical sensor, we also investigate
the ability to discriminate between a finite set of possible indenter
shapes. We do not discard the possibility of learning more contact
parameters in the future.

In the following sections, we present concrete implementations of
these concepts using two transduction methods, based on resistance and
optics respectively. For each prototype (one using resistance and two
using optics) we present our manufacturing methods, data collection
protocol, and results analysis.

\section{Resistivity based sensor}

When applying the spatially overlapping signals method with
piezoresistance as the underlying transduction method, the sensor
consists of a single continuous volume of piezoresistive polymer with
a number of embedded electrodes (Fig.~\ref{fig:three_sensors}, Left).

\subsection{Sensor manufacturing}

To achieve piezoresistance for our silicone, we disperse multiwall
carbon nanotubes (MWCNT, purity: 85\%, Nanolab Inc.) into PDMS
(Sylgard 184, Dow Corning). The key aspect for this procedure is
choosing the appropriate ratio of conductive filler to
elastomer. According to the percolation theory, a composite will
display the most pronounced piezoresistive effect for a ratio referred
to as the percolation threshold by Hu et al.~\cite{hu2011}. In order to find this
value for our materials, we tested a series of samples with the
concentrations of MWCNTs from 0.2wt.\% to 5wt.\%. We found that the
most pronounced change in conductivity occurred around the threshold
of 0.4wt.\% filler, which we used in all subsequent experiments.

In order to achieve uniform distribution of carbon nanotubes within
PDMS, we use chloroform as a common solvent, an approach referred to
as the solution casting method by Liu and Choi~\cite{liu2012}. First, we combine
chloroform and MWCNT and use a horn-type ultrasonicator in a pulse
mode with 50\% amplitude for 30 min to evenly disperse the
MWCNTs. Then we add PDMS to the mix (at chloroform:PDMS weight ratio
of 6:1 or more to reduce the viscosity of the mixture), stir for 5 min
to diffuse the PDMS, and then sonicate again for 30 min. We then heat
the mixture at 80$^{\circ}$C for 24 hours to evaporate the
chloroform. After adding the curing agent, the mixture is ready to be
poured into the mold; for the experiments presented here we
empirically selected a curing agent to PDMS ratio of 1:20. Finally,
the sample is cured in an oven at 80$^{\circ}$ C for 4 hours.

In order to isolate piezoresistive effects from mechanical changes at
the contacts due to indentation, we mechanically separated the wire
contacts from the piezoresistive sample placed under indentation
tests. We extended a number of 30mm side channels from the sample,
each filled with a CNT-filled PDMS mixture with a higher concentration
of 2.5wt.\% We then embedded copper wires directly into the mixture at
the end of these channels (Fig.~\ref{fig:three_sensors}, Left). The mixture with
the ratio of 2.5wt.\% has no piezoresistive characteristics and its
conductivity is close to that of the copper wires; thus, the mixture
with the ratio of 0.4wt.\% located at the center of the mold dominates
the overall conductivity.

Our sampling circuit measures the change in resistance between all
pairs of terminals that occurs as a result of some strain being
applied to our sample material. For each one of our six terminal
pairs, we take a baseline resistance measurement. This value is then
compared against the real time measurement after strain is applied
using an instrumentation amplifier. The amplified signal is directly
fed to an analog to digital converter. A switching matrix scheme
allows us to quickly perform this procedure on all six sensing
pairs. The overall circuit delivers the set of all six measurements
every 25 milliseconds, resulting in a 40Hz sampling frequency.

\subsection{Data collection}

We collect training and test data with a planar stage (Marzhauser
LStep) and a linear probe located above the stage to indent vertically
on the sensor with a 6mm hemispherical tip. The probe is
position-controlled and the reference level is set manually such that
the tip barely makes contact with the sensor. The probe does not have
force sensing capability, hence we use indentation depth as a proxy
for indentation force.

For indentation locations, we use two patterns. The \textit{grid
  indentation pattern} consists of a regular 2D grid of indentation
locations, spaced 2mm apart along each axis. However, the order in
which grid locations are indented is randomized. This is in contrast
with the \textit{random indentation pattern}, where the locations of
the indentations are sampled randomly over the surface of the
sample. The grid indentation pattern is used for training our
algorithms, while the random indentation pattern is used as a test
set.

Taking into account the tip diameter, plus a 1mm margin such that we
do not indent directly next to an edge, our regular indentation
pattern results in 54 indent locations distributed over a 10mm by 16mm
area (9x6 grid).

For each indentation location, we sample the signal from each pair of
electrodes at a depth of 3mm. Each such measurement $i$ results in a tuple
of the form $\Phi_i=(x_i,y_i,d_i,r_i^1,..,r_i^6)$, where $x_i$ and
$y_i$ represent the location of the indentation, $d_i$ is the
indentation depth, and $r_i^1,..,r_i^6$ (also referred to collectively
as $\bm{r}_i$) represent the change in the six resistance values we
measure between depth $d_i$ and depth 0 (the probe on the surface of
the sample). These tuples are used for data analysis as described in
the next section.

\subsection{Analysis and results}

Our first goal is to learn the mapping from all terminal pairs
readings $\bm{r}_i$ to the indentation location $(x,y)$. To train the
predictor, we collected four data sets in regular grid patterns,
totaling 216 indentations (9x6 grid, hence each set contains 54
indentations) . For testing, we collected a dataset consisting of 60
indentations in a random pattern. All indentations were performed to a
depth of 3mm, or 50\% of the total depth of the sample. The metric
used to quantify the success of this mapping is the magnitude of the
error (in mm) between the predicted indentation position and ground
truth. In our analysis below, we report this error for individual test
points, as well as its mean, median and standard deviation over the
complete testing set.

The baseline that we compare against includes a ``Center
Predictor'' and a ``Random Predictor''. The former always predicts
the location of the indentation on the center of our sample, and the
later predicts a completely random location within the sample
surface. The useful area of our sample is 16mm by 10mm; on our test
set, the Center Predictor produces a median error of 5mm, while the
random predictor, if given a large test set, converges on a median
error of above 6mm.

We first attempted Linear Regression as our learning method. The
results were significantly better than the baseline, with a median
error of under 2mm. Still, visual inspection of the magnitude and
direction of the errors revealed a circular bias towards the center
that we attempted to compensate for with a different choice of
learning algorithm. The second regression algorithm we tested was
Ridge Regression with a Laplacian kernel. The Laplacian kernel is a
simple variation of the ubiquitous radial basis kernel, which explains
its ability to remove non-linear bias. In this case, we used the first
half of the training data for training the predictor, and the second
half to calibrate the ridge regression tuning factor $\lambda$ and the
kernel bandwidth $\sigma$ through grid search.

\begin{figure}[t]
\centering
\includegraphics[clip, trim=0.0cm 0.6cm 0.0cm 2.0cm, width=0.40\textwidth]{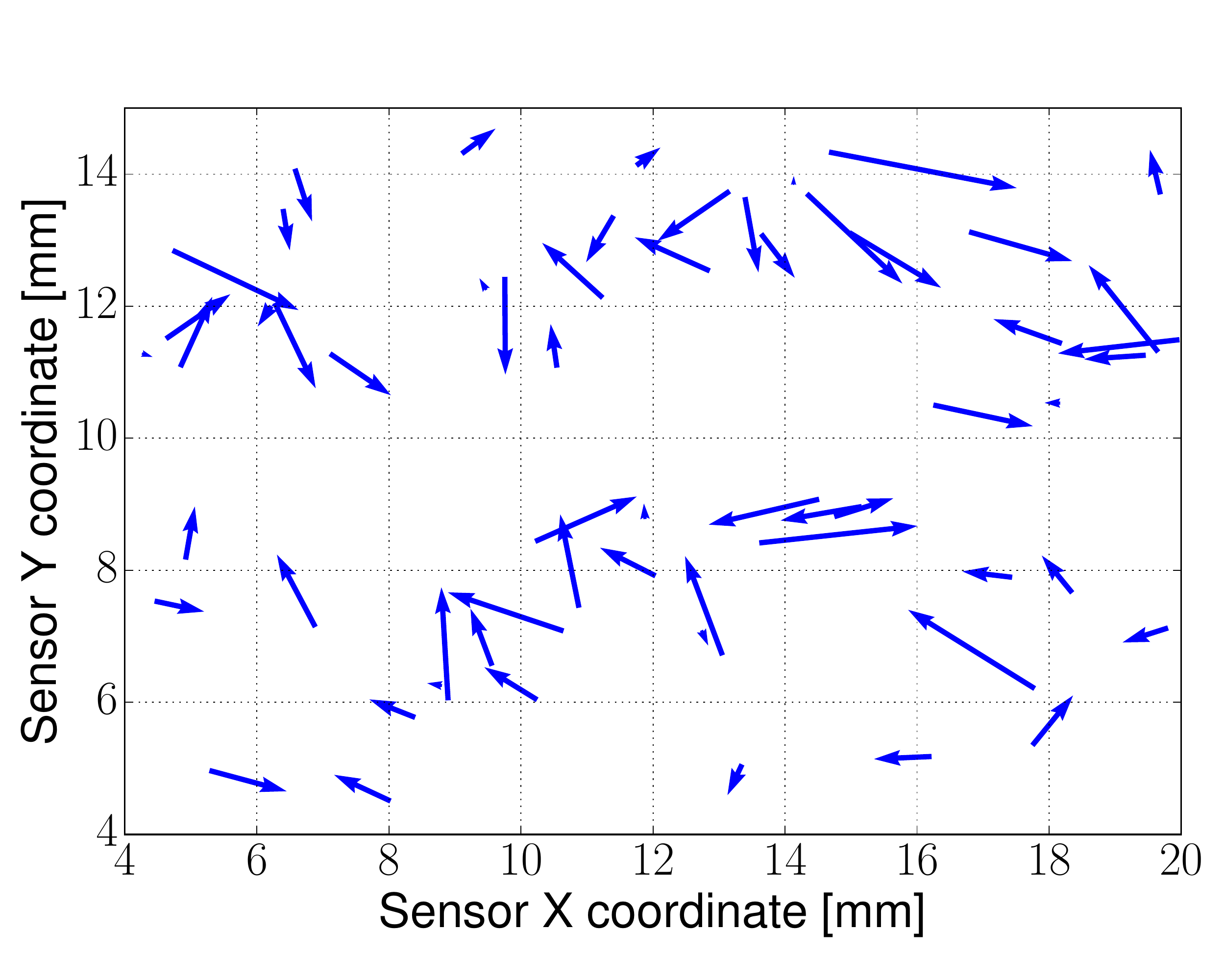}
\caption{Magnitude and direction of localization errors for resistive
  sensor. Each arrow represents one test indentation: the base of the
  arrow is at the ground truth location while the tip is at the
  predicted location.}
\label{fig:vector}
\vspace{-0mm}
\end{figure}

The numerical results using both of our predictors, as well as the two
baseline predictors, are summarized in Table~\ref{table1}. These
results are aggregated over the complete test set consisting of 60
indentations. Linear regression identifies the location of the
indentation within 2mm on average, while Laplacian ridge regression
($\lambda=2.7e^{-2}$, $\sigma=6.15e^{-4}$) improves these
results to sub-millimeter median accuracy. In addition to the
aggregate results, Fig.~\ref{fig:vector} illustrates the magnitude and
direction of the localization error for the ridge regression.

\begin{table}[b]
\centering
\caption{Indentation location prediction accuracy, resistive sensor}
\label{table1}
\begin{tabular}{lccc}
\hline
\\[-3mm]
\multicolumn{1}{l}{\small \textbf{Predictor}} &\small \textbf{Median Err.} &\small \textbf{Mean Err.} &\small \textbf{Std. Dev.} \\ \hline
\\[-3mm] \hline
\\[-2mm]
\small Center predictor                & \small 5.00 mm      & \small 5.13 mm    & \small 2.00 mm        \\ 
\small Random predictor                & \small 6.30 mm      & \small 6.70 mm    & \small 3.80 mm        \\ 
\small Linear regression               & \small 1.75 mm      & \small 1.75 mm    & \small 0.83 mm        \\ 
\small \textbf{Lapl. ridge regr.}      & \small \textbf{0.97 mm}      & \small \textbf{1.09 mm}    & \small \textbf{0.59 mm}        \\ 
\end{tabular}
\end{table}

\subsection{Discussion and limitations}

Our results with the piezoresistive sensor illustrate both the
advantages and the downsides of this transduction method. Considering
first the positive features, we note that any terminal can act as both
emitter and receiver, and any terminal requires a single wire. As such,
this method has the potential to provide a very rich data set with
very few wires. This is illustrated by our sensor, which achieves
sub-millimeter localization accuracy over a $400 mm^2$ area with only
four total wires.

\begin{figure}[t]
\setlength{\tabcolsep}{0mm}
\begin{tabular}{cc}
\includegraphics[clip, trim=0.0cm 0.0cm 0.0cm 0.0cm, width=0.48\columnwidth]{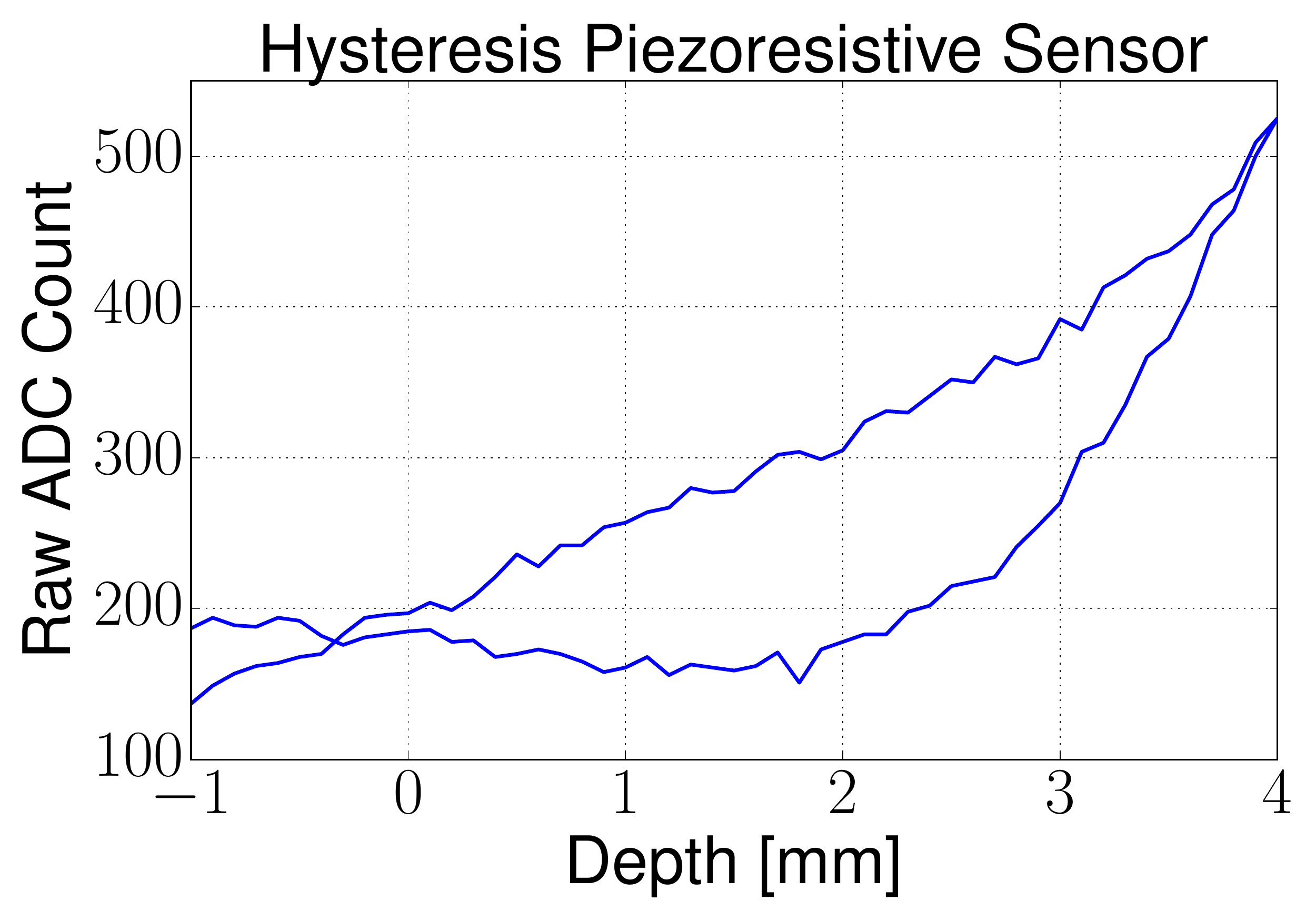}&
\includegraphics[clip, trim=0.0cm 0.0cm 0.0cm 0.0cm, width=0.48\columnwidth]{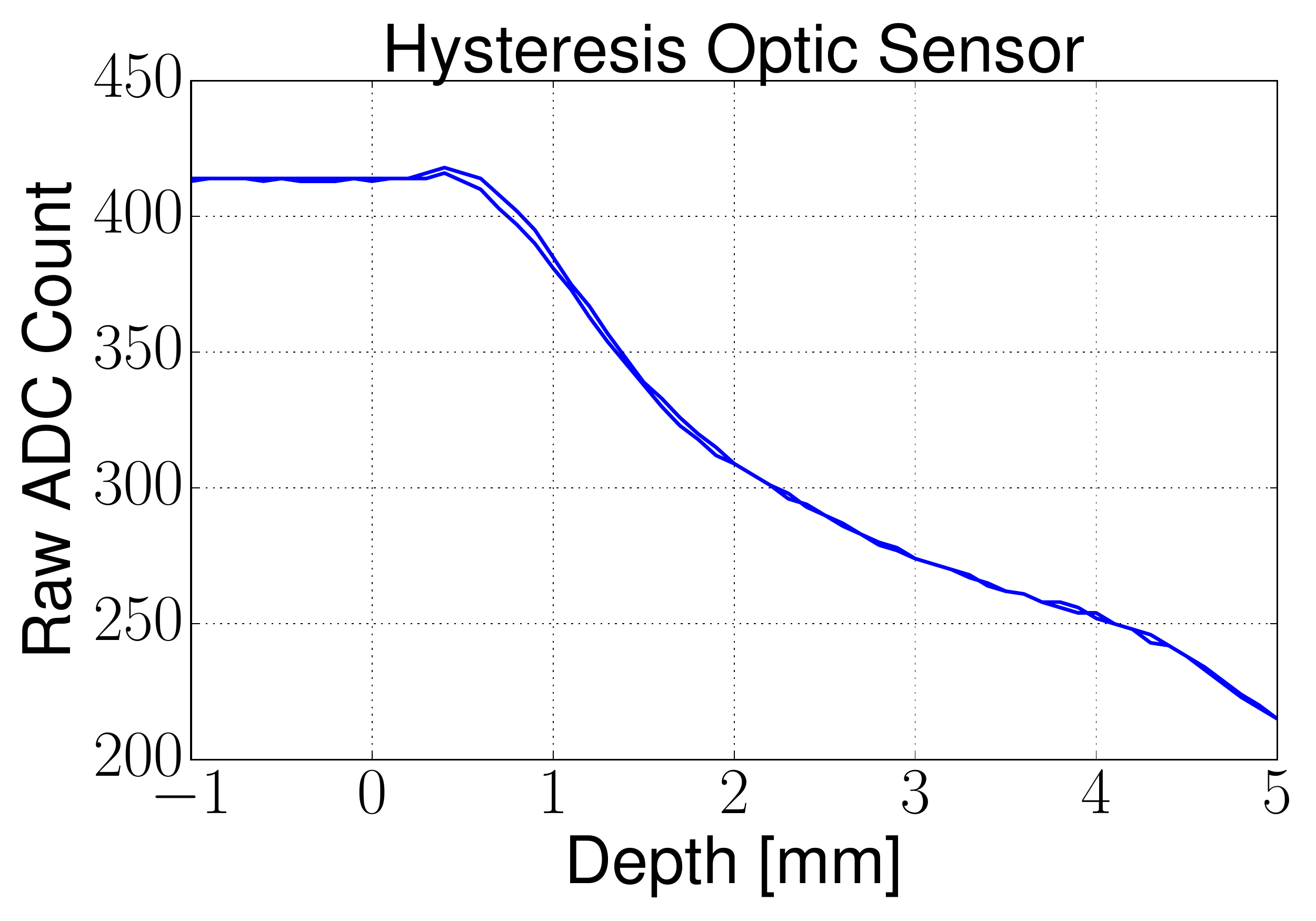}\\[-2mm]
\end{tabular}
\caption{Hysteresis graph for piezoresistive and optic sensors. Graphs show a single terminal pair response to an indentation
up to 4mm and 5mm respectively. A hemispherical indenter was used in both cases.}
\label{fig:hysteresis}
\end{figure}

However, we have also found a number of limitations that could not be
avoided using our manufacturing techniques. The mechanical interface
between the terminals and the piezoresistive elastomer significantly
affects resistance measurements; in this proof-of-concept prototype,
we used side channels with highly conductive elastomer to isolate the
sensor, but this complicates future integration in robot fingers. We
also found that the sensor exhibits significant hysteresis over time
scales on the order of seconds, as well as signal drift over longer
time scales (days). Fig.~\ref{fig:hysteresis}, Left shows the resulting
signal from a particular terminal pair during a controlled indentation 
event. We mitigate this here by using the resistance
values at depth 0 (just before touching) as a baseline that is
subtracted from all signals collected during that indentation, and by
always indenting to a known depth, but these methods again limit
future applicability to robot hands.

We believe that overlapping piezoresistive signals are worth
considering, as they hold significant promise for collecting rich data
with very few terminals, with additional improvements needed in the
manufacture of piezoresistive elastomers with embedded electrodes. In
this study, we now move to a different transduction method based on
optics, where we trade off an increase in wire count (and thus
complexity) for a much simpler manufacturing method and improved SNR 
which yields greatly improved performance and the ability to predict depth.

\section{Optics based sensor}
\label{sec:optics}

When using optics as a transduction method, our sensor consists of a
continuous volume of optically clear elastomer, with embedded light
emitters and receivers. The first version we describe has emitters and
receivers only along the perimeter of the sensing area
(Fig.~\ref{fig:three_sensors}, Middle); in the next section we will
consider a version with terminals also embedded on the base.

An important difference compared to the piezoresistive sensor is the
fact that now our sensing terminals have defined roles as either an
emitter (LED) or a receiver (photodiode). The all-pairs approach still
applies; we pair each emitter with all possible receivers. Note that
in this paper we will refer to both an LED or photodiode as a
``sensing terminal''.

\subsection{Light transport and interaction modes}
\label{sec:modes}

\begin{figure*}[t]
\setlength{\tabcolsep}{1mm}
\centering
\begin{tabular}{ccc}
\raisebox{28mm}{
\begin{tabular}{c}
\includegraphics[clip, trim=1.5cm 3cm 1.5cm 3.0cm, width=0.225\linewidth]{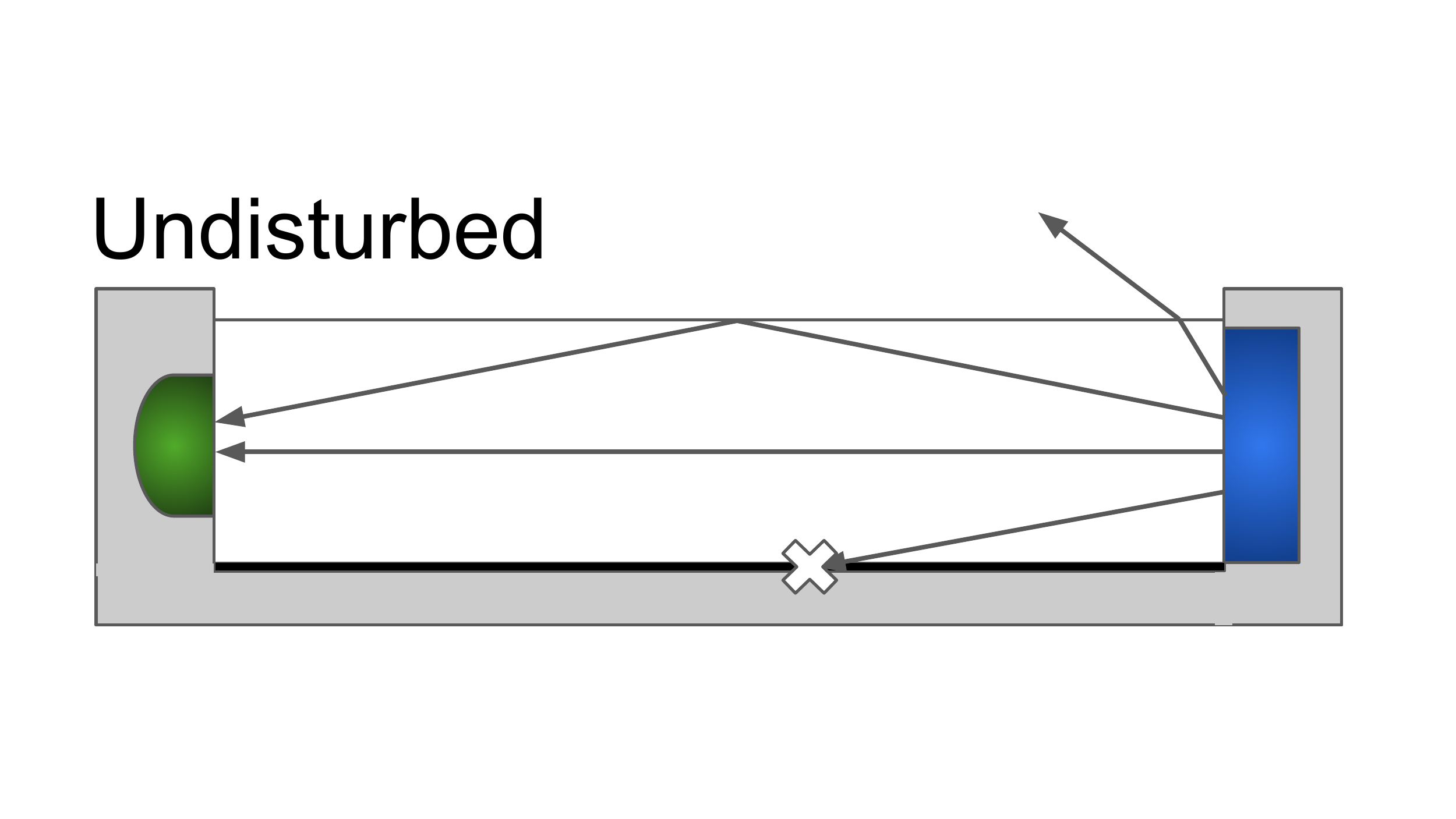}\\[1mm]
\includegraphics[clip, trim=1.5cm 3cm 1.5cm 3.0cm, width=0.225\linewidth]{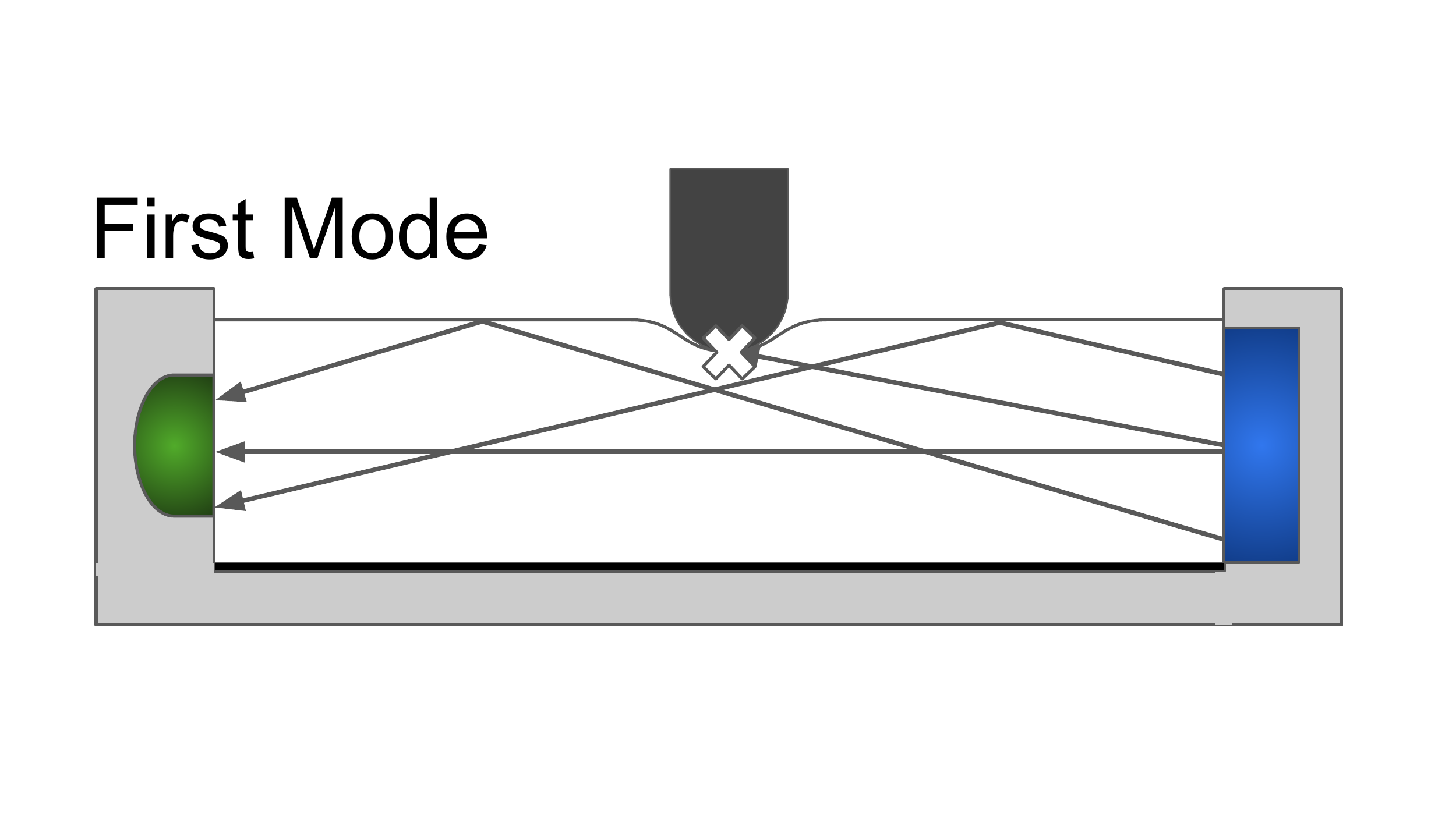}\\[1mm]
\includegraphics[clip, trim=1.5cm 3cm 1.5cm 3.0cm, width=0.225\linewidth]{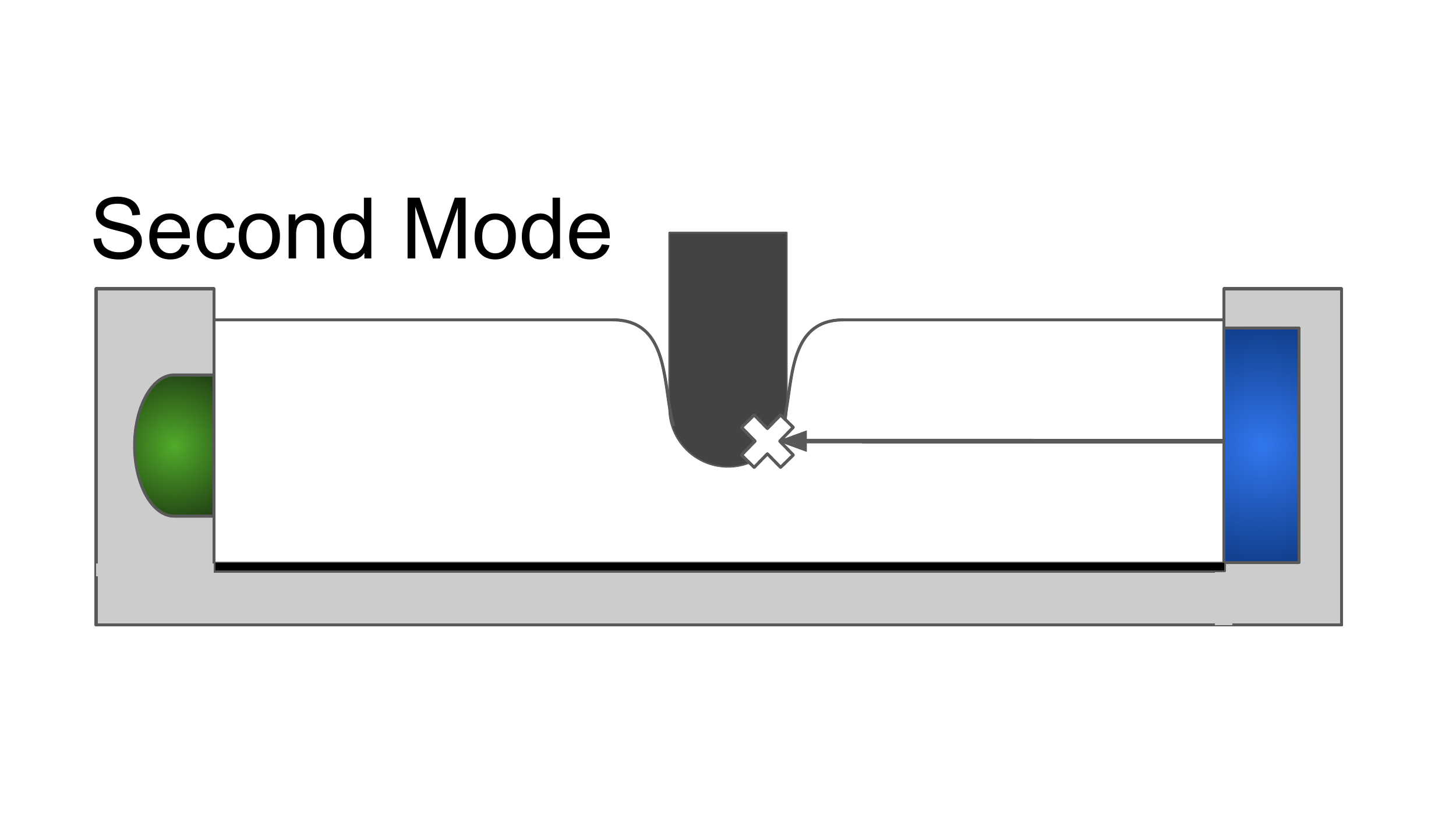}\\[1mm]
\end{tabular}
}
&
\raisebox{9mm}{
\includegraphics[clip, trim=0.4cm 0.5cm 0.5cm 0.5cm, width=0.3\linewidth]{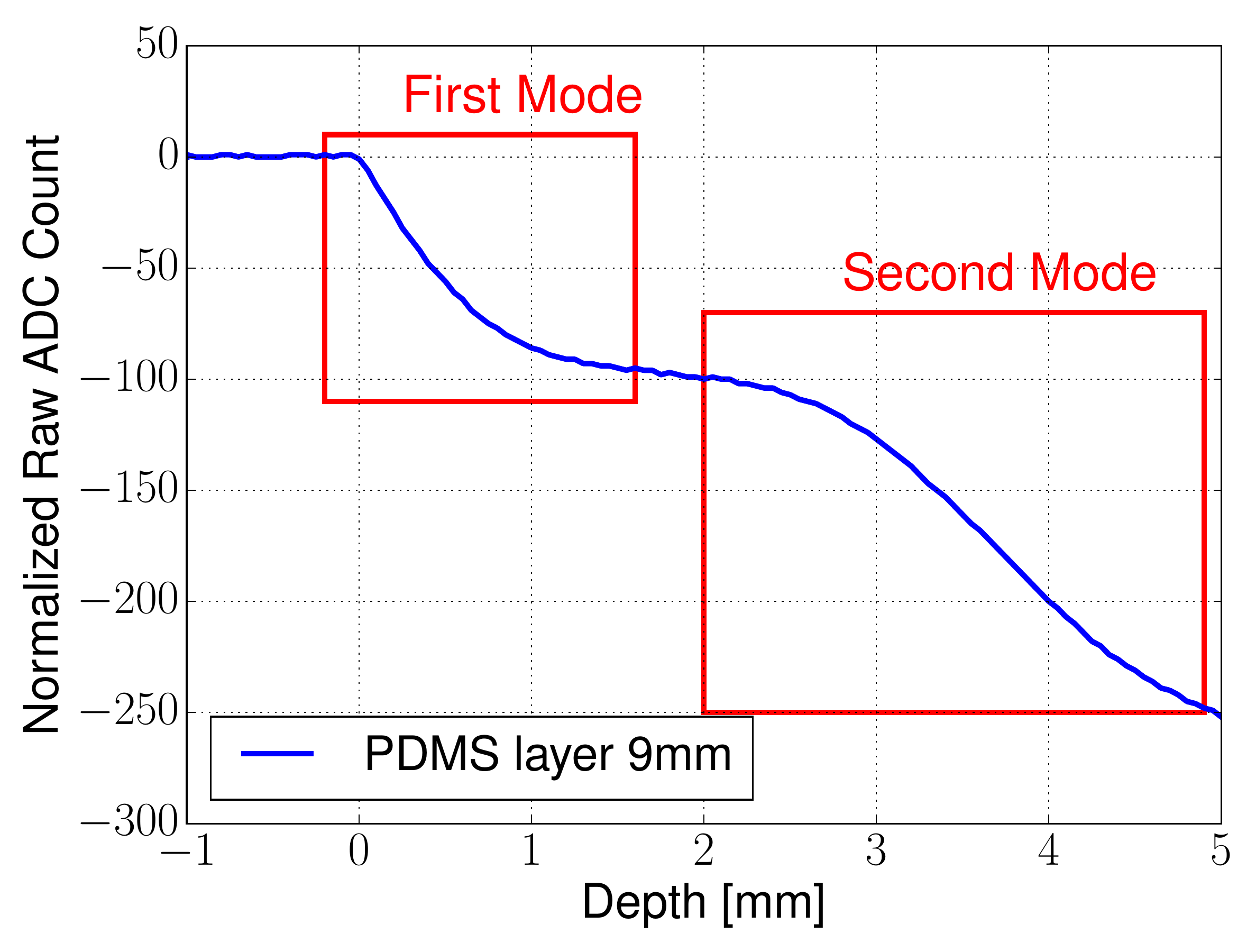}
}
&
\raisebox{4mm}{\includegraphics[clip, trim=0.4cm 0.5cm 0.5cm 0.5cm, width=0.38\linewidth]{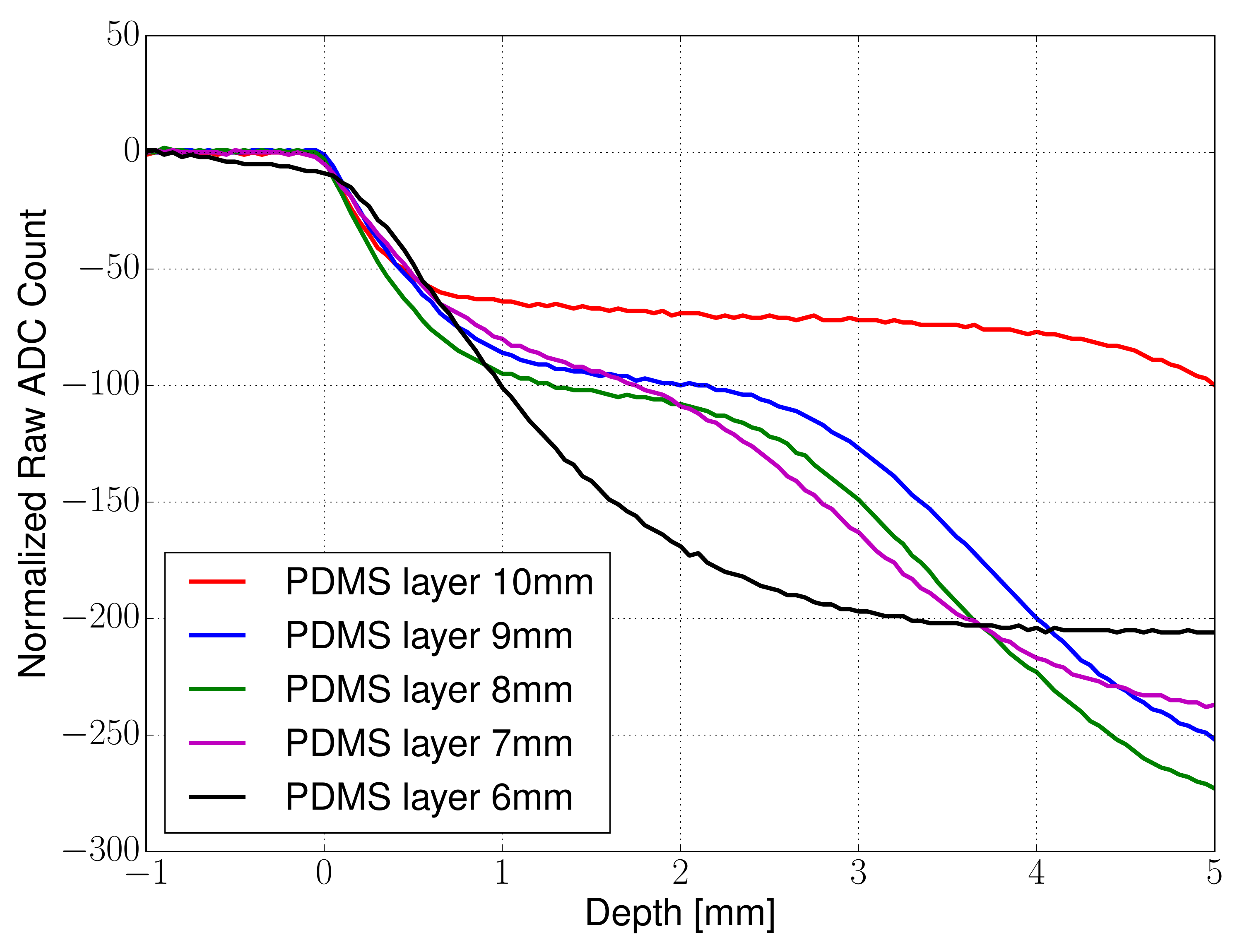}
}
\end{tabular}
\caption{\textbf{Left and middle:} interaction modes. The first mode of detection
  happens upon light contact and manifests as a sudden drop in the
  signal. The second mode is activated with a heavier contact, in this
  particular case after a depth of approximately 2mm. \textbf{Right:}
  we investigated this behavior for different heights of the PDMS
  layer, in order to select one for which these two modes are
  contiguous.}
\label{fig:Signal_modes}
\end{figure*}

We first focus on a single emitter-receiver pair in order to discuss
the underlying light transport mechanism in detail. The core
transduction mechanism relies on the fact that as a probe indents the
surface of the sensor, light transport between the emitter and the
receiver is altered, changing the signal reported by the receiver
(Fig.~\ref{fig:Signal_modes}). Consider the multiple ways in which
light from the LED can reach the opposite photodiode: through a direct
path or through a reflection. In particular, based on Snell's law, due
to different refractive indices of the elastomer and air, light rays
hitting the surface below the critical angle are reflected back into
the elastomer.

As the probe makes initial contact with the sensor surface, the
elastomer-air interface is removed from the contact area and surface
normals are immediately disturbed. This changes the amount of light
that can reach the diode via surface reflection. \textit{This is the
  first mode of interaction that our transduction method captures.} It
is highly sensitive to initial contact, and very little penetration
depth produces a strong output signal.

As the depth of indentation increases, the indenter starts to also
block light rays that were reaching the photodiode through a
direct, line-of-sight path. \textit{This is our second mode of
  interaction.} To produce a strong signal, the probe must reach deep
enough under the surface where it blocks a significant part of the
diode's surface from the LED's vantage point.

We note that other light paths are also possible between the emitter
and receiver. The interface between the clear elastomer and the
holding structure (the bottom and side walls of the cavity) can also
give rise to reflections. While we do not explicitly consider their
effects, they can still produce meaningful signals that are captured
by the data-driven mapping algorithm.

We would like our sensor to take advantage of both operating modes
described above, noting that one is highly sensitive to small
indentations while the other provides a strong response to deeper
probes. \textit{We thus aim to design our sensor such that these two
  modes are contiguous as the indentation depth increases.} The goal
is to obtain high sensitivity throughout the operating range of the
sensor. The key geometric factor affecting this behavior is the
height of the elastomer layer, which we determine experimentally.

To investigate this factor, we constructed multiple 3D printed square
molds with LEDs (SunLED XSCWD23MB) placed 20mm in opposition from the
photodiode (Osram SFH 206K), with PDMS filling the cavity. Work by
Johnson and Adelson~\cite{johnson2009} shows PDMS and air have approximate refractive
indexes of 1.4 and 1.0 respectively, with a critical angle of $45$
degrees. Each mold was filled with PDMS to achieve a different layer thickness
and then indented to 5mm depth in a midpoint location between the 
LED and the photodiode. Results are shown in Fig.~\ref{fig:Signal_modes}.
We note that prototypes with PDMS layers over 10mm exhibited a dead
band: after a certain threshold depth the photodiode signal does not
change as we indent further down until you indent deep enough to
activate the second mode. In contrast, the 7mm layer provides the best
continuity between our two modes, while the 8mm layer gives good
continuity while also producing a stronger signal when indented. We
thus built all our subsequent sensors with an 8mm PDMS layer.

\subsection{Sensor manufacture}

We validate this concept on a sensor comprised of 8 LEDs and 8
photodiodes arranged in an alternating pattern and mounted in sockets
along the central cavity walls. For this sensor we used through-hole 
technology (THT) components for ease of prototyping. We used a 5mm x 5mm x 7.2mm sized LED model 
SunLED XSCWD23MB with peak wavelength 460nm and a Osram SFH 206K 
photodiode with peak sensitivity at 850nm and a 2.65mm x 2.65mm sensitive area.
To build the sensor we use a 3D printed square mold with exterior dimensions of 
48mm x 48mm. The cavity in the mold is 32mm x 32mm. Any LED is thus 
able to excite multiple photodiodes.

As this sensor only has terminals around its perimeter, the base
simply consists of 3D printed ABS plastic. However, we found that the
clear elastomer and the holding plastic exhibit bonding/unbonding
effects at a time scale of 5-10s when indented, creating unwanted
hysteresis. To eliminate such effects, we coat the bottom of the
sensor with a 1mm layer of elastomer saturated with carbon black
particles (shown in Fig.~\ref{fig:Signal_modes} by a thick black
line). This eliminates bottom surface reflections, and exhibits no
adverse effects, as the clear elastomer permanently bonds with the
carbon black-filled layer. As shown in Fig.~\ref{fig:hysteresis}, this
sensor exhibits little to no hysteresis between an indentation and the
subsequent retraction, in contrast to the resistive version.

An Arduino Mega 2560 handles LED switching and taking analogue
readings of each photodiode. The photodiode signal is amplified
through a standard trans-impedance amplifier circuit, and each LED is
driven at full power using an NPN bipolar junction transistor. The
resulting sampling frequency with this setup is 60Hz.

\subsection{Data collection}
\label{sec:optics_collection}

When collecting data, we read signals from all the photodiodes as
different LEDs turn on. Having 8 LEDs gives rise to 8 signals for each
photodiode, plus an additional signal with all LEDs turned off for a
total of 9 signals per diode. This last signal allows us to measure
the ambient light captured by each diode and subtract it from all
other signals, so that the sensor can perform consistently in
different lighting situations.

The main difference compared to the piezoresistive sensor is that we
treat indentation depth as an additional variable. At each location we
use the following protocol. We consider the sensor surface to be the
reference level, with positive depth values corresponding to the
indenter tip going deeper into the sensor. We collect data at both
negative and positive depths. For depths between $-10mm$ and $-1mm$,
we collect one data point every $1mm$. The indenter then goes down to
a depth of $5.0mm$ taking measurements every $0.1mm$. The same
procedure is mirrored with the indenter tip retracting.

Each measurement $i$ results in a tuple of the form
$\Phi_i=(x_i,y_i,d_i,p_{j=1}^1,..,p_{j=1}^8,...,p_{j=9}^1,..,p_{j=9}^8)$
where $(x_i,y_i)$ is the indentation location in sensor coordinates,
$d_i$ is the depth at which the measurement was taken and
$(p_{j}^1,..,p_{j}^8)$ correspond to readings of our 8 photodiodes as
we turn each LED on: state $j\ \epsilon\ [1,8]$, from which we
subtract the ambient light captured by each diode when all LEDs are
turned off (state $j=9$). We thus have a total of 75 numbers in each
tuple $\Phi_i$. 

Similar to the previous sensor, we use a \textit{grid indentation
  pattern} for training data and a \textit{random indentation pattern}
for testing. Taking into account the diameter of our tip, plus a 3mm
margin such that we do not indent directly next to an edge, our grid
indentation pattern results in 121 locations distributed over a 20mm x
20mm area (11x11 grid). We allow for a bigger margin compared to the
resistive sensor since the components are mounted directly on the
walls of our mold.

\subsection{Analysis and results}
\label{sec:optics_results}

Our objective here is to learn the mapping between our photodiode
readings $(p_{j}^1,..,p_{j}^8)$ for $j\ \epsilon\ [1,8]$ to the
indentation location and depth $(x_i,y_i,d_i)$. Note that the ambient light measurement
(state $j=9$) is not used as a feature, since it has already been subtracted from
other states. 

To train our predictors we collected four \textit{grid pattern}
datasets, each consisting of 121 indentations, and each indentation
containing 161 datapoints at different depths. Aiming for robustness
to changes in lighting conditions, two of these datasets were
collected with the sensor exposed to ambient light and the other two
datasets were collected in darkness. The feature space used for
training has a dimensionality of 64. For testing, we collected two
datasets using the \textit{random indentation pattern}, one collected
in ambient light and another collected in the dark, with 100
indentation events in each.

\begin{figure}[t]
\centering 
\includegraphics[clip, trim=0.4cm 0.4cm 0.4cm 0.4cm, width=0.85\linewidth] {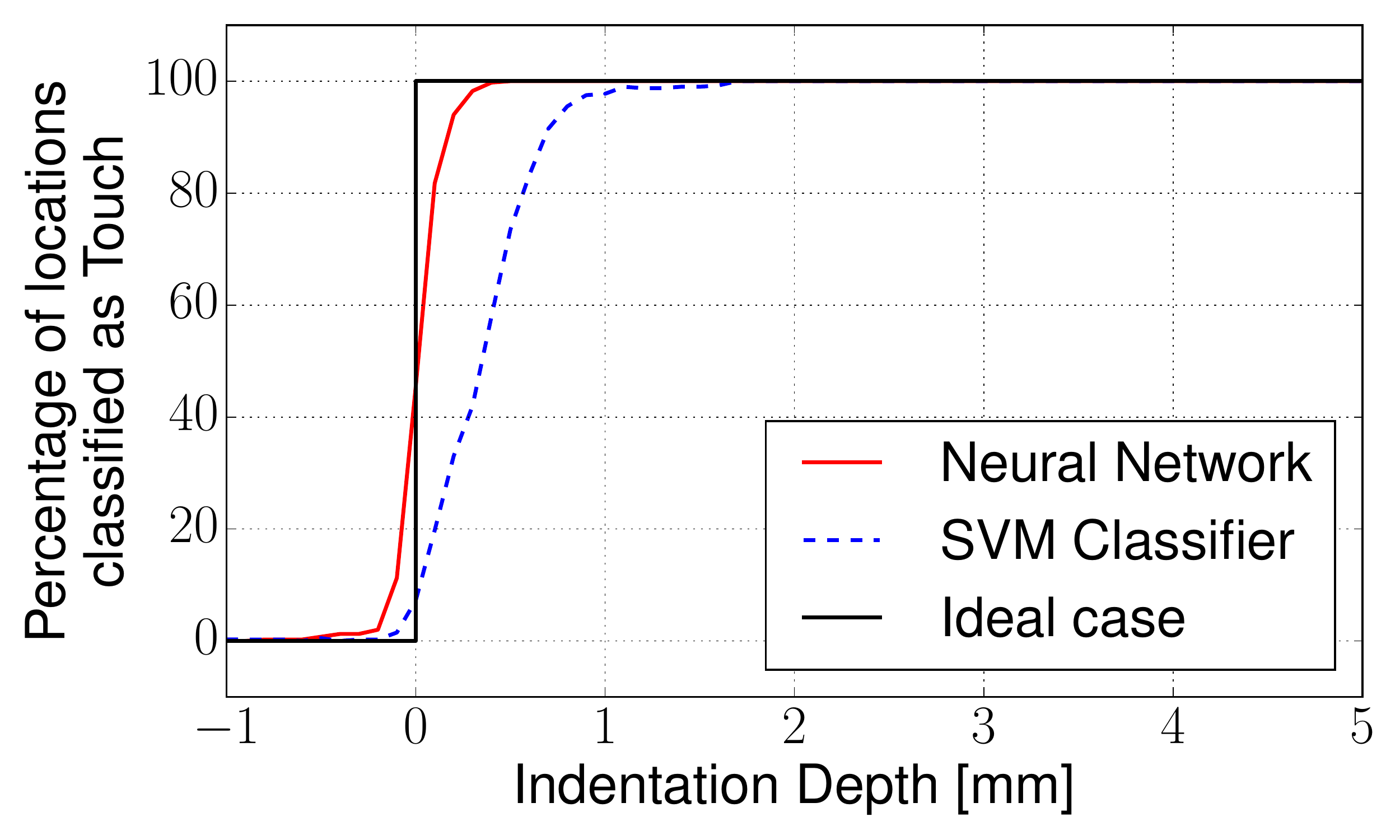}
\caption{Touch classification results over all locations in our test
  dataset. Values in the graph represent the percentage of locations
  in our test dataset where touch was predicted for an indentation at
  a given depth.}
\label{fig:Classifier_graph}
\end{figure}

\mystep{Touch classification.} The first step we take is to use a
classifier to determine if touch is occurring; this classifier is
trained on both data points with $d_i < 0$ and $d_i \geq 0$. We tested
two different classification methods: a linear SVM and a Neural
Network (NN) with one input layer, one hidden layer (1024 nodes, ReLU
activation function) and one output layer (1 node, sigmoid activation
function). Results for both classifiers are presented in
Fig.~\ref{fig:Classifier_graph}. For any given depth, we show the
percentage of indentations (over all of our test locations) that are
classified as touch.

For an indentation depth of 0.1 mm the NN classifier correctly
classifies 82\% of the events as ``touch'', increasing to 94\% at 0.2
mm and hitting 98\% at 0.3 mm. The price we pay is an 11\%
misclassification rate at -0.1 mm depth (slightly above the
surface). The SVM classifier rarely interprets cases above the surface
as touch, but starts reliably predicting touch (above 80\% of cases)
at a depth of 0.5 mm or higher, hitting 99\% after 1 mm. Overall,
using the NN classifier, and according to the elasticity profile shown
in Fig.~\ref{fig:Mapping}, we detect touch with 82\% reliability for a
force of 0.11 N (0.1 mm indentation) and 98\% reliability at 0.55 N
(0.3 mm) or higher.

\mystep{Depth and location regression}. If the classifier predicts
touch is occurring, we use a second stage regressor to predict values
for $(x_i,y_i,d_i)$. This regressor is trained only on training data
with $d_i \geq 0$. We use a kernelized ridge regressor with a
Laplacian kernel and use half of the training data to calibrate the
ridge regression tuning factor $\lambda$ and the kernel bandwidth
$\sigma$ through grid search. However, the computational requirements
of training a ridge regressor with a non-linear kernel require downsampling of
our training data: for each indentation, we only trained on
indentation depths in 0.5 mm increments starting from the surface of
the sensors. Results presented in this section were obtained with
$\lambda = 2.15e^-4$ and $\sigma = 5.45e^-4$.

\begin{table}[t]
\small
\centering
\caption{Localization and depth accuracy}
\label{light_table}
\begin{tabular}{cc}
\textbf{Ambient Light Dataset}\\
\hspace{-6mm}
\setlength{\tabcolsep}{1.5mm}
\begin{tabular}{ccccccc}
\hline
\\[-3mm]
\multicolumn{1}{c|}{\textbf{Depth}} & \multicolumn{3}{|c}{\textbf{Localization Error} (mm)}                      & \multicolumn{3}{|c}{\textbf{Depth Error} (mm)}                             \\
\multicolumn{1}{c|}{(mm)}               & \multicolumn{1}{c}{\textbf{Median}} &  \multicolumn{1}{c}{\textbf{Mean}} & \multicolumn{1}{c|}{\textbf{Std. Dev}}  & \multicolumn{1}{c}{\textbf{Median}} &  \multicolumn{1}{c}{\textbf{Mean}} & \multicolumn{1}{c}{\textbf{Std. Dev}} \\ \hline
\\[-3mm] \hline
\\[-2mm]
0.1          & 2.18                & 2.36              & 1.50              & 0.16                & 0.16              & 0.07              \\
0.5          & 0.71                & 0.83              & 0.61              & 0.05                & 0.06              & 0.04              \\
1.0          & 0.64                & 0.74              & 0.53              & 0.04                & 0.06              & 0.05              \\
2.0          & 0.42                & 0.50              & 0.31              & 0.04                & 0.04              & 0.03              \\
3.0          & 0.36                & 0.41              & 0.25              & 0.03                & 0.04              & 0.04              \\
5.0          & 0.31                & 0.35              & 0.22              & 0.08                & 0.09              & 0.06             
\end{tabular}
\\[6mm]
\\
\textbf{Dark Dataset}\\
\hspace{-6mm}
\setlength{\tabcolsep}{1.5mm}
\begin{tabular}{ccccccc}
\hline
\\[-3mm]
\multicolumn{1}{c|}{\textbf{Depth}} & \multicolumn{3}{|c}{\textbf{Localization Error} (mm)}                      & \multicolumn{3}{|c}{\textbf{Depth Error} (mm)}                             \\
\multicolumn{1}{c|}{(mm)}               & \multicolumn{1}{c}{\textbf{Median}} &  \multicolumn{1}{c}{\textbf{Mean}} & \multicolumn{1}{c|}{\textbf{Std. Dev}}  & \multicolumn{1}{c}{\textbf{Median}} &  \multicolumn{1}{c}{\textbf{Mean}} & \multicolumn{1}{c}{\textbf{Std. Dev}} \\ \hline
\\[-3mm] \hline
\\[-2mm]
0.1          & 3.50                & 3.83              & 2.20              & 0.18                & 0.18              & 0.07              \\
0.5          & 1.10                & 1.33              & 0.98              & 0.05                & 0.05              & 0.04              \\
1.0          & 0.77                & 0.84              & 0.53              & 0.05                & 0.06              & 0.04              \\
2.0          & 0.54                & 0.65              & 0.45              & 0.04                & 0.05              & 0.04              \\
3.0          & 0.41                & 0.48              & 0.32              & 0.03                & 0.05              & 0.04              \\
5.0          & 0.31                & 0.39              & 0.28              & 0.08                & 0.10              & 0.07             
\end{tabular}
\end{tabular}
\end{table}

\begin{figure}[t]
\setlength{\tabcolsep}{0mm}
\begin{tabular}{ccc}
\includegraphics[clip, trim=1.70cm 0.0cm 1.0cm 0.0cm, width=0.35\columnwidth]{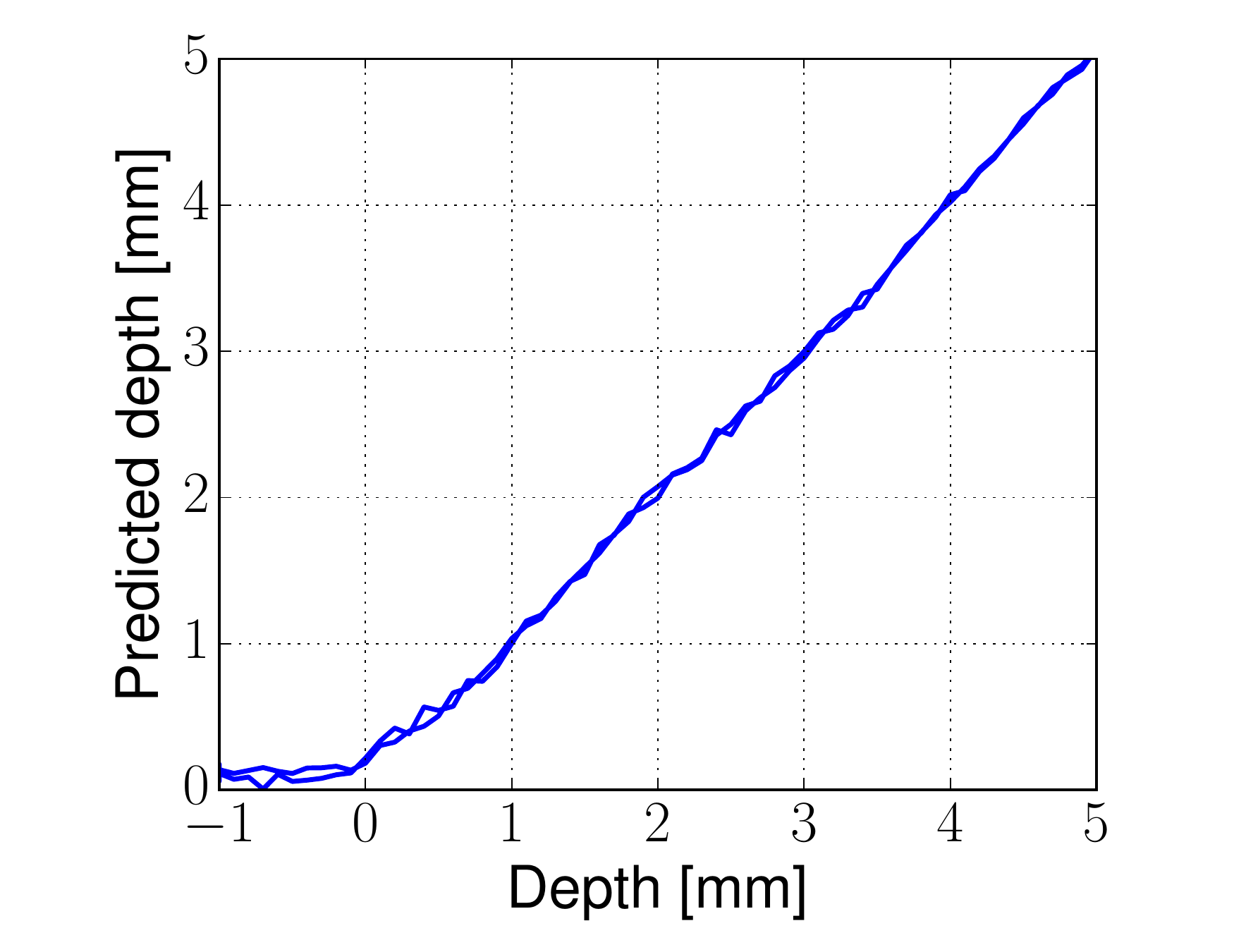}&
\includegraphics[clip, trim=1.65cm 0.0cm 1.0cm 0.0cm, width=0.35\columnwidth]{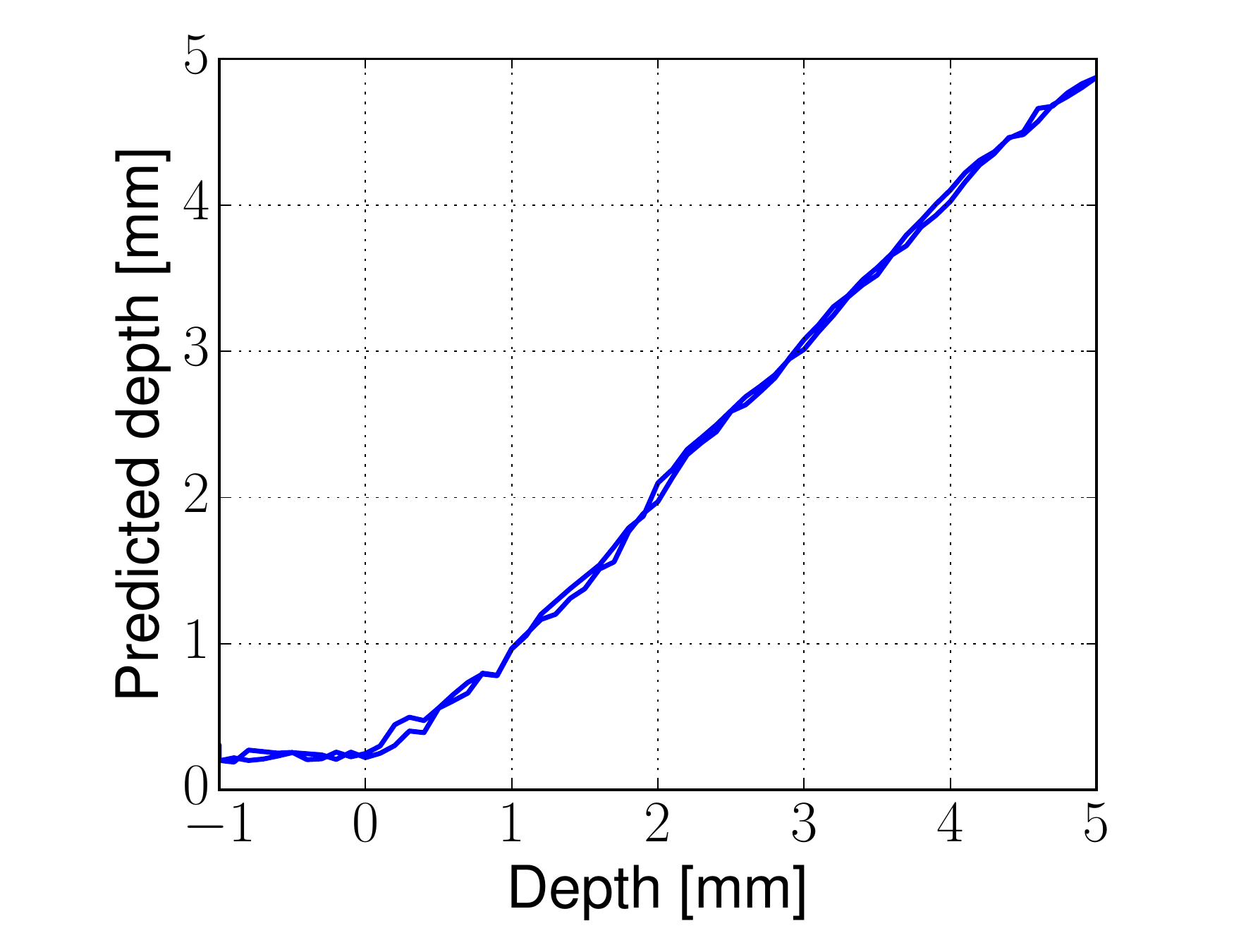}&
\includegraphics[clip, trim=1.65cm 0.0cm 1.0cm 0.0cm, width=0.35\columnwidth]{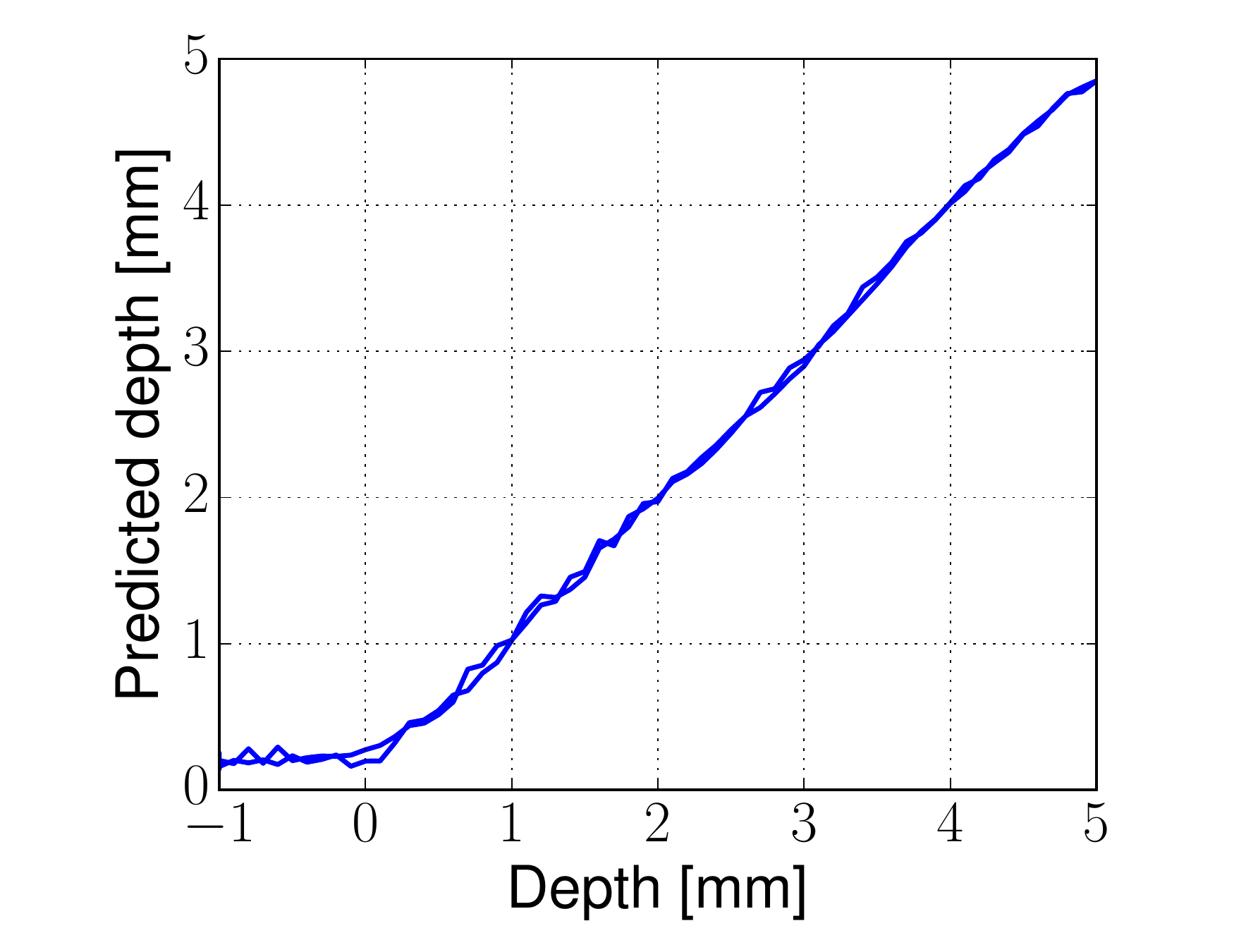}\\[-2mm]
\footnotesize Center&
\footnotesize Edge&
\footnotesize Corner
\end{tabular}
\caption{ Regression results for depth prediction at three of the
  random locations in our test datases, close to the center (sensor
  coordinates 16.5,19.1), an edge (6.4, 18.3) and a corner (10.9,
  6.2). }
\label{fig:depth}
\end{figure}

\begin{figure}[t]
\setlength{\tabcolsep}{1mm}
\begin{tabular}{cc}
\includegraphics[width=0.49\columnwidth]{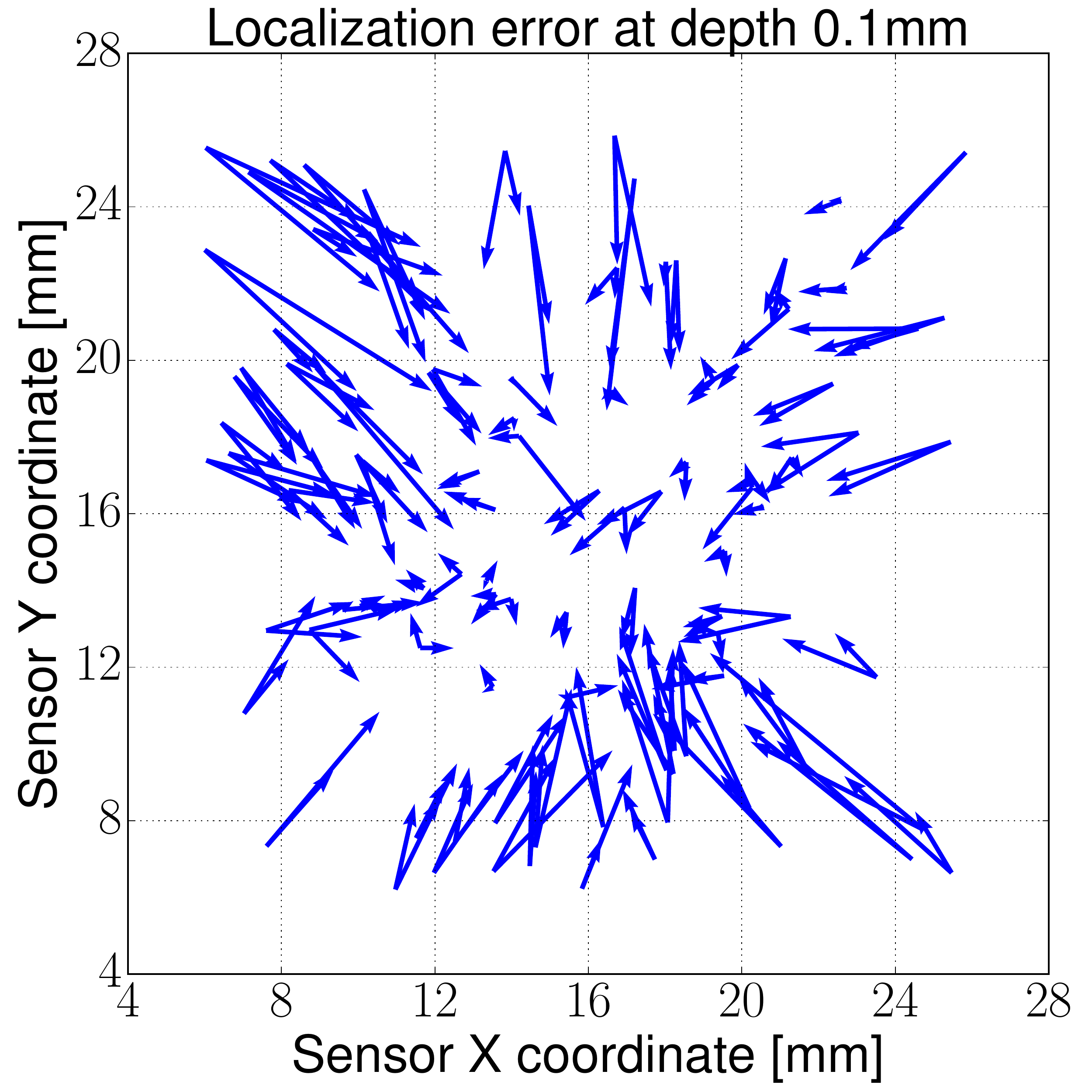}&
\includegraphics[width=0.49\columnwidth]{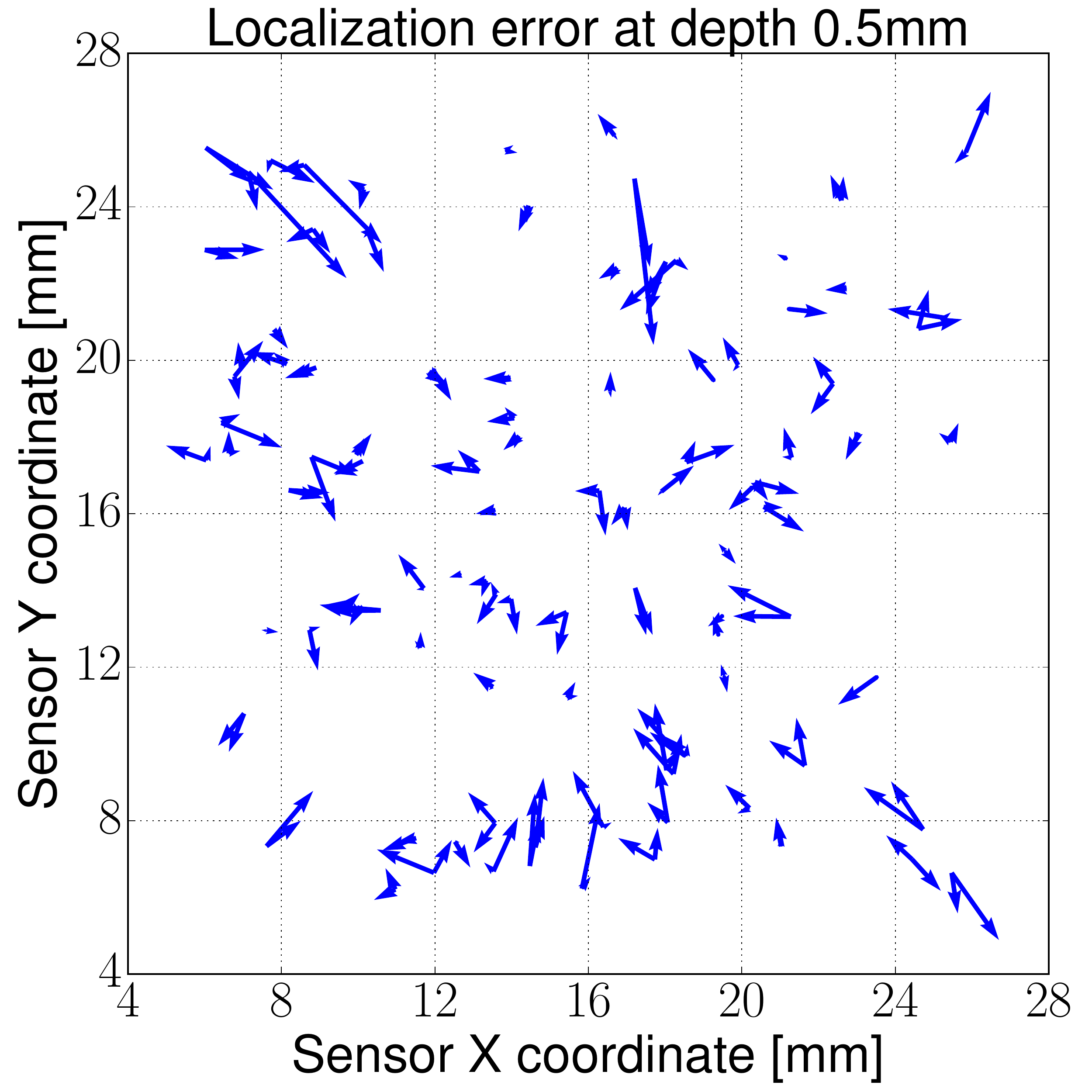}\\[1mm]
\includegraphics[width=0.49\columnwidth]{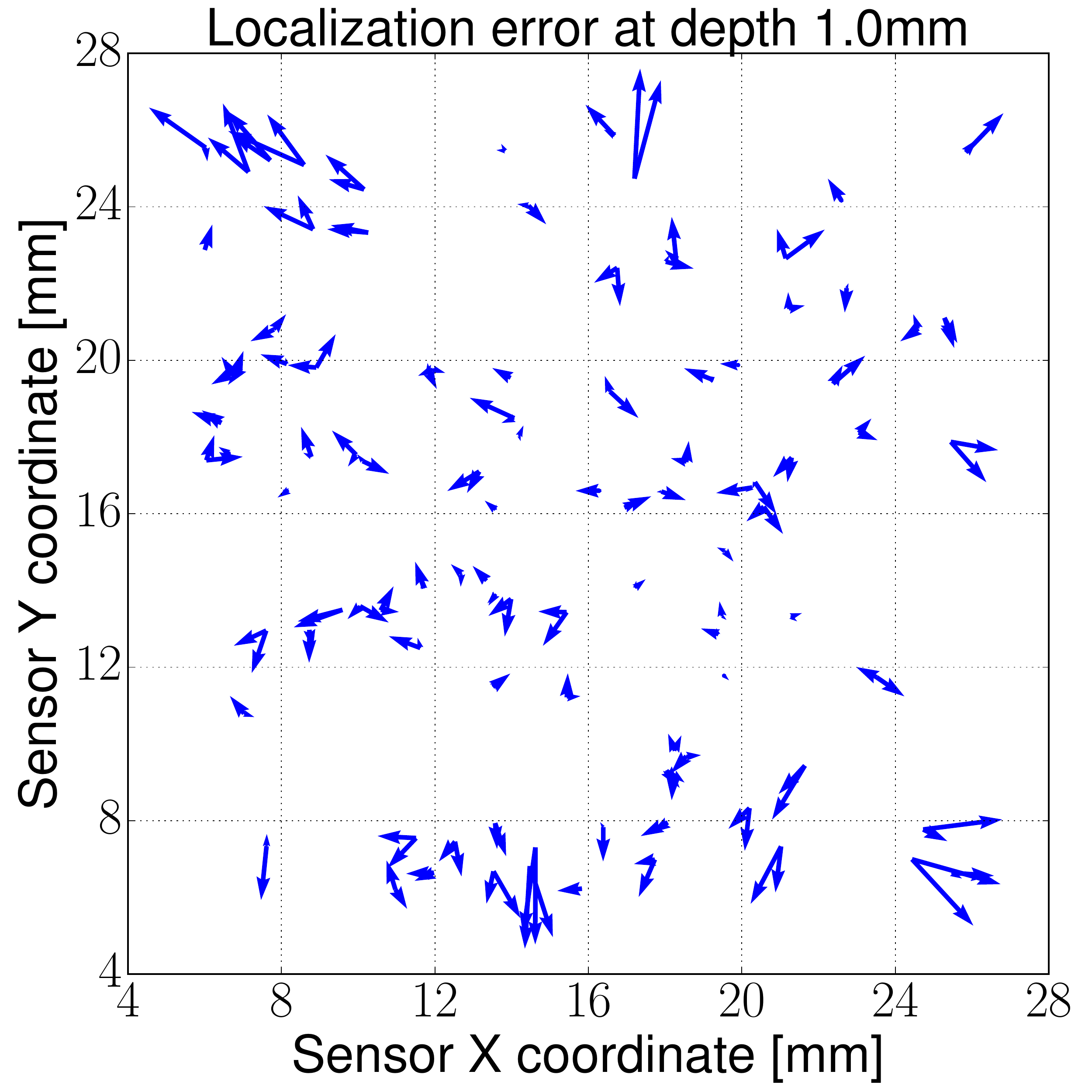}&
\includegraphics[width=0.49\columnwidth]{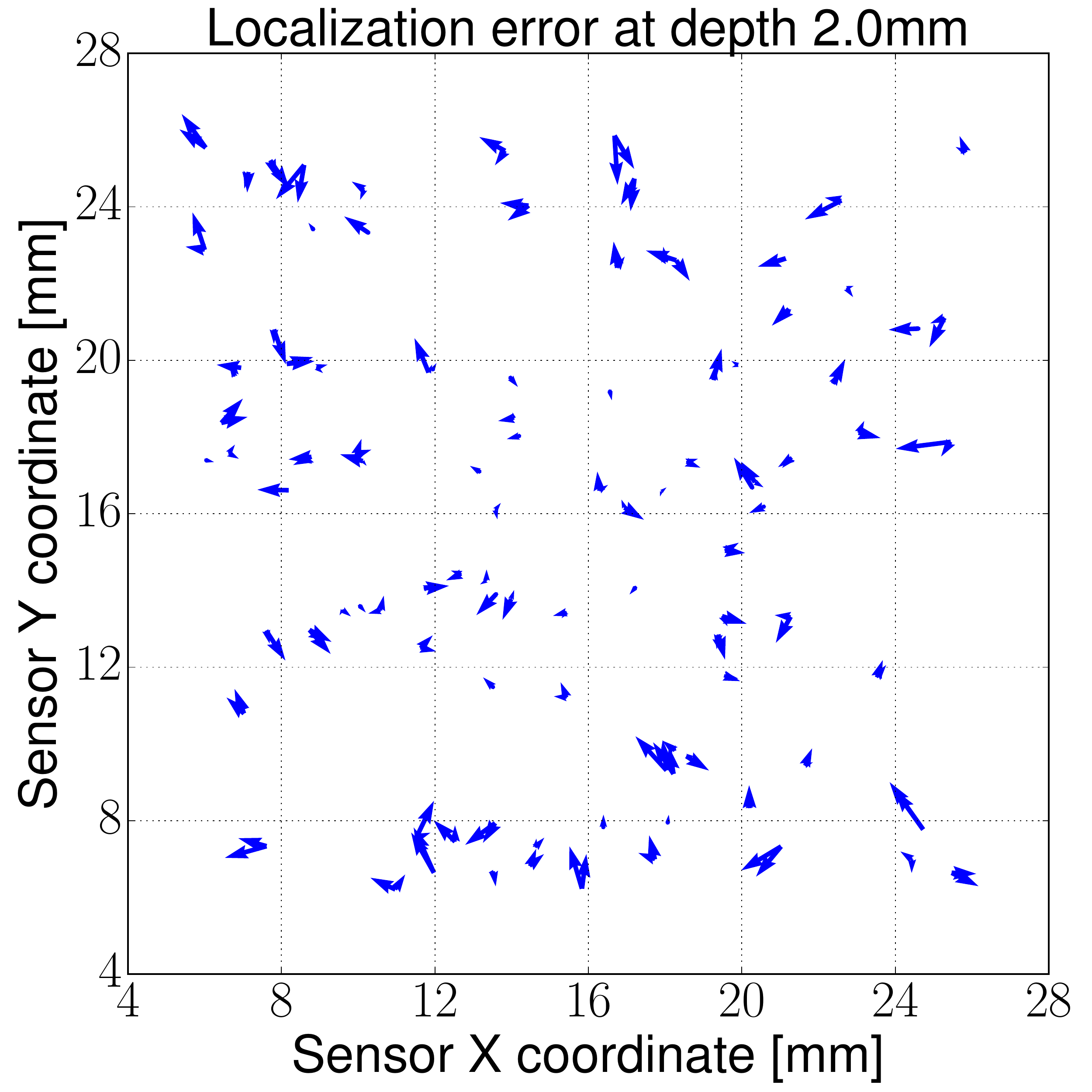}\\[1mm]
\includegraphics[width=0.49\columnwidth]{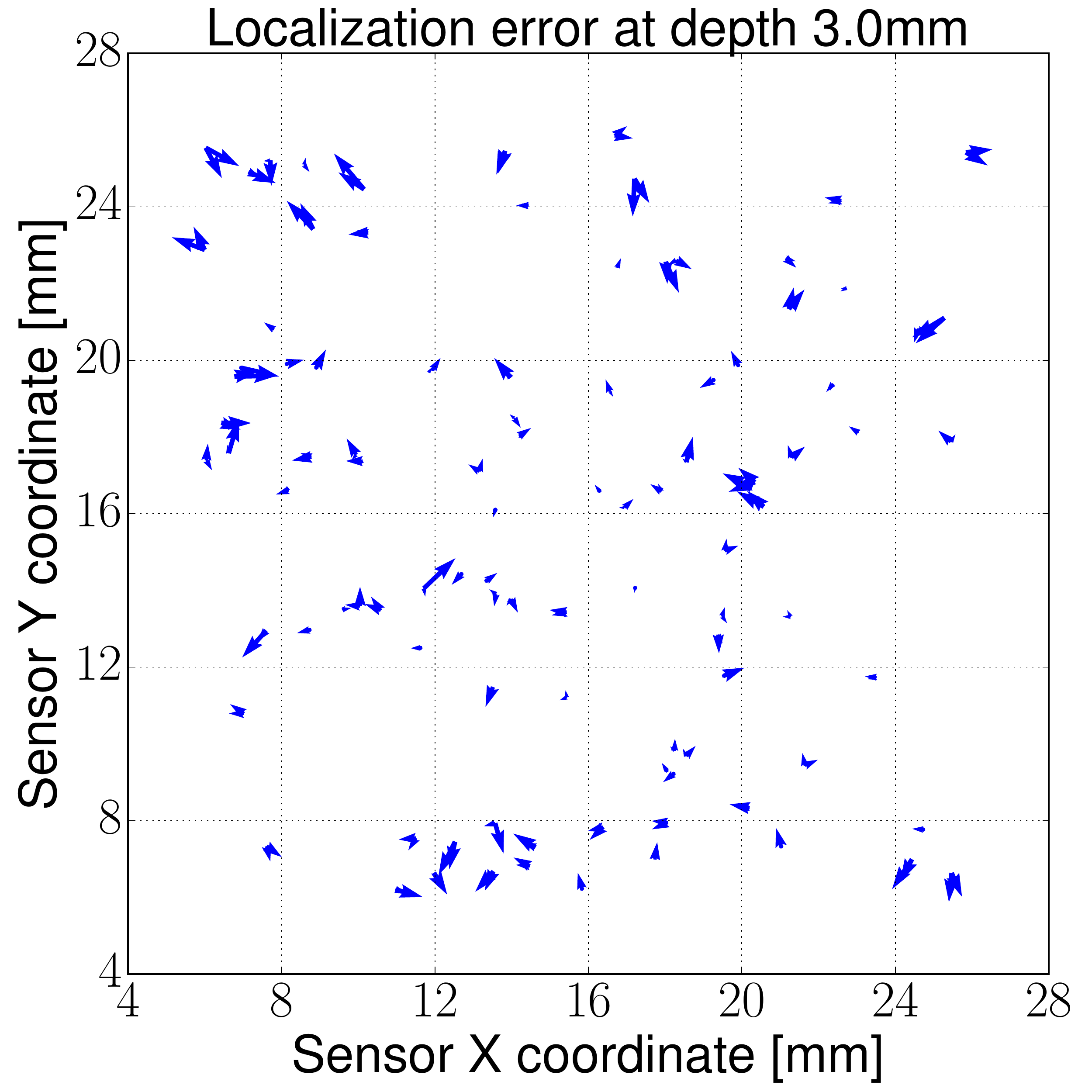}&
\includegraphics[width=0.49\columnwidth]{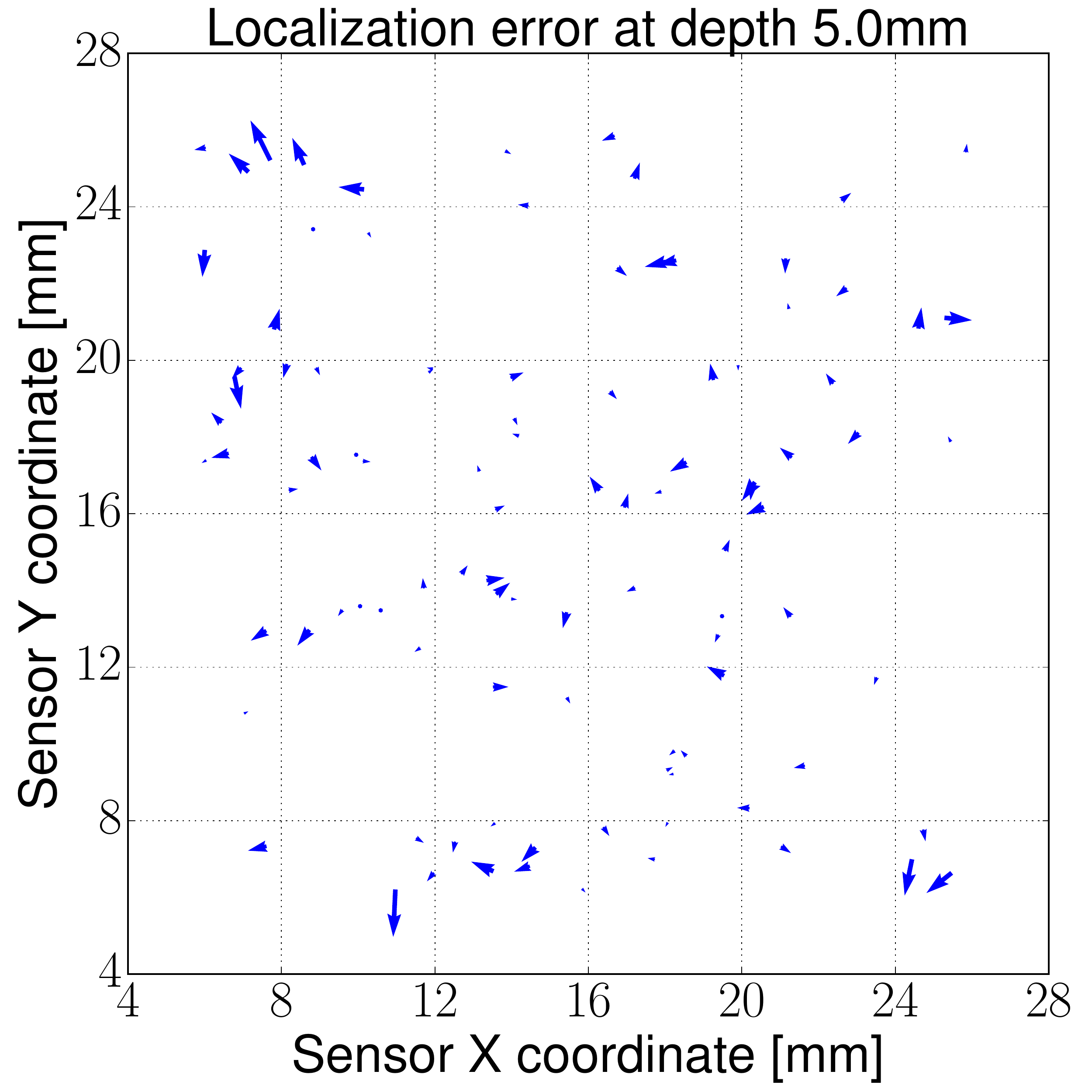}
\end{tabular}
\caption{Localization results for ambient light test dataset. Each
  arrow represents one indentation in our test set; the base is at the
  ground truth location while the tip of the arrow shows the predicted
  location.}
\label{fig:Light_results}
\end{figure}

The metric used to quantify the success of our regressor is the
magnitude of the error for both the localization and depth
accuracy. Detailed numerical results for both of these metrics, and
for both ambient light and dark datasets, are presented in
Table~\ref{light_table}, aggregated over our entire test sets.

To provide more insight into the behavior of the sensor, we present
additional visualizations for the error in both depth and
location. The results for depth can only be visualized for specific
locations. Fig.~\ref{fig:depth} shows the performance at three
different locations in our ambient light test dataset. Since this
regressor is trained only on contact data, it is not able to predict
negative depths which causes error to be greater at depths close to
0mm. At 0.5mm and deeper, the depth prediction shows an accuracy of
below half a millimeter.

Localization results on the light dataset can be visualized in
Fig.~\ref{fig:Light_results} for a few representative depths. At depth
0.1mm the signals are still not good enough to provide accurate
localization, but as we indent further down localization improves well
beyond sub-millimeter accuracy.

\mystep{Signal removal analysis.} Given the nature of our spatially
overlapping signals method, it is of special interest to analyze the
relationship between the number of sensing terminals and the resulting
accuracy in touch localization and depth prediction. Our hypothesis is
that a higher number of sensing terminals will yield higher accuracy,
but how we distribute these terminals to over the intended sensing
area also plays an important role in determining performance.

We consider four different cases for our optical sensor. In each case,
we discard the signals from a subset of terminals of the base
configuration (which has a total of 16 terminals). We compare these
cases between themselves, and against the base configuration. We use
the localization and depth error at a depth of 2mm for the sake of
readability (other depths follow the same trend).

\begin{figure}[t]
\setlength{\tabcolsep}{0mm}
\begin{tabular}{cc}
\includegraphics[width=0.5\columnwidth]{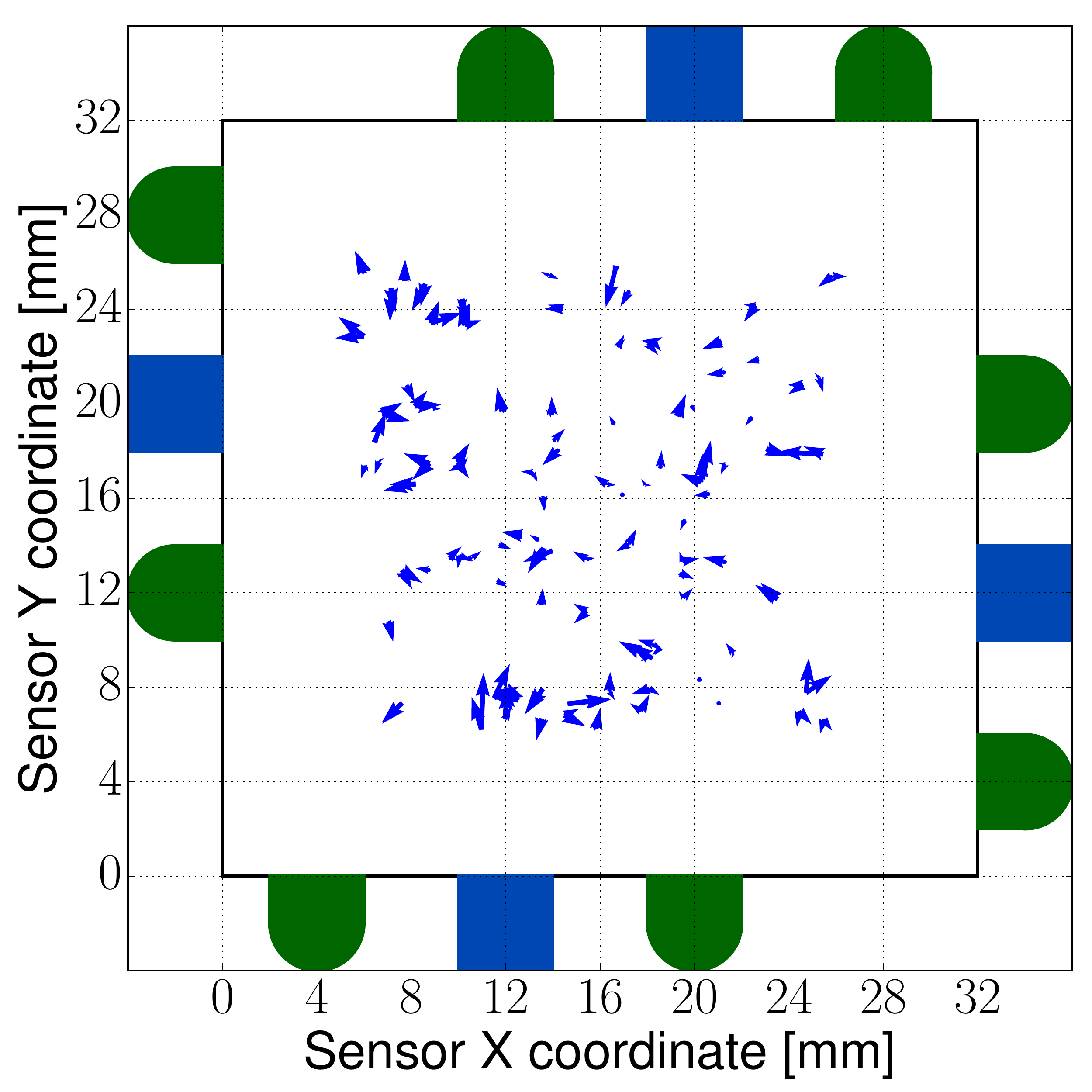}&
\includegraphics[width=0.5\columnwidth]{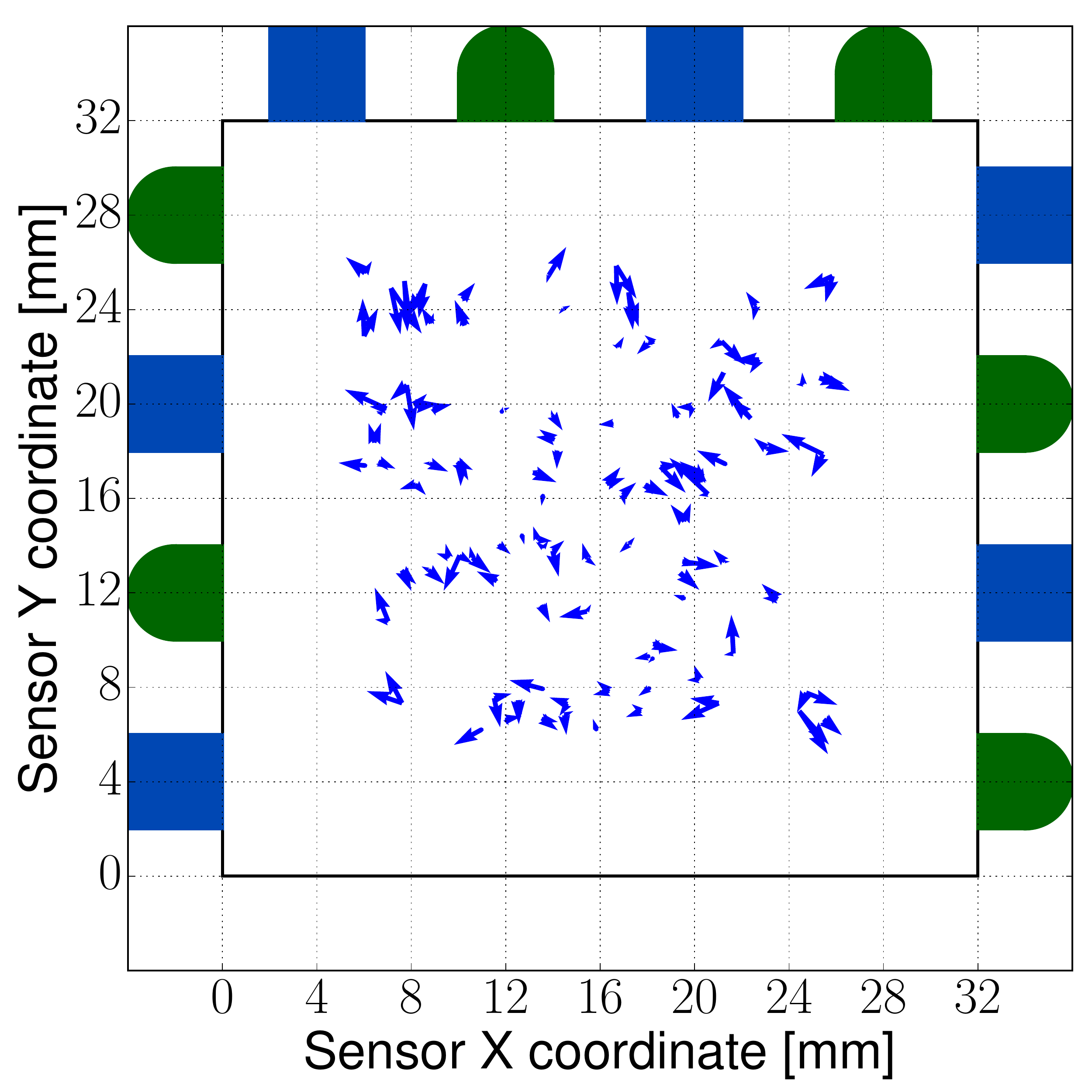}\\[-2mm]
\footnotesize (a) Case 1&
\footnotesize (b) Case 2\\[2mm]
\includegraphics[width=0.5\columnwidth]{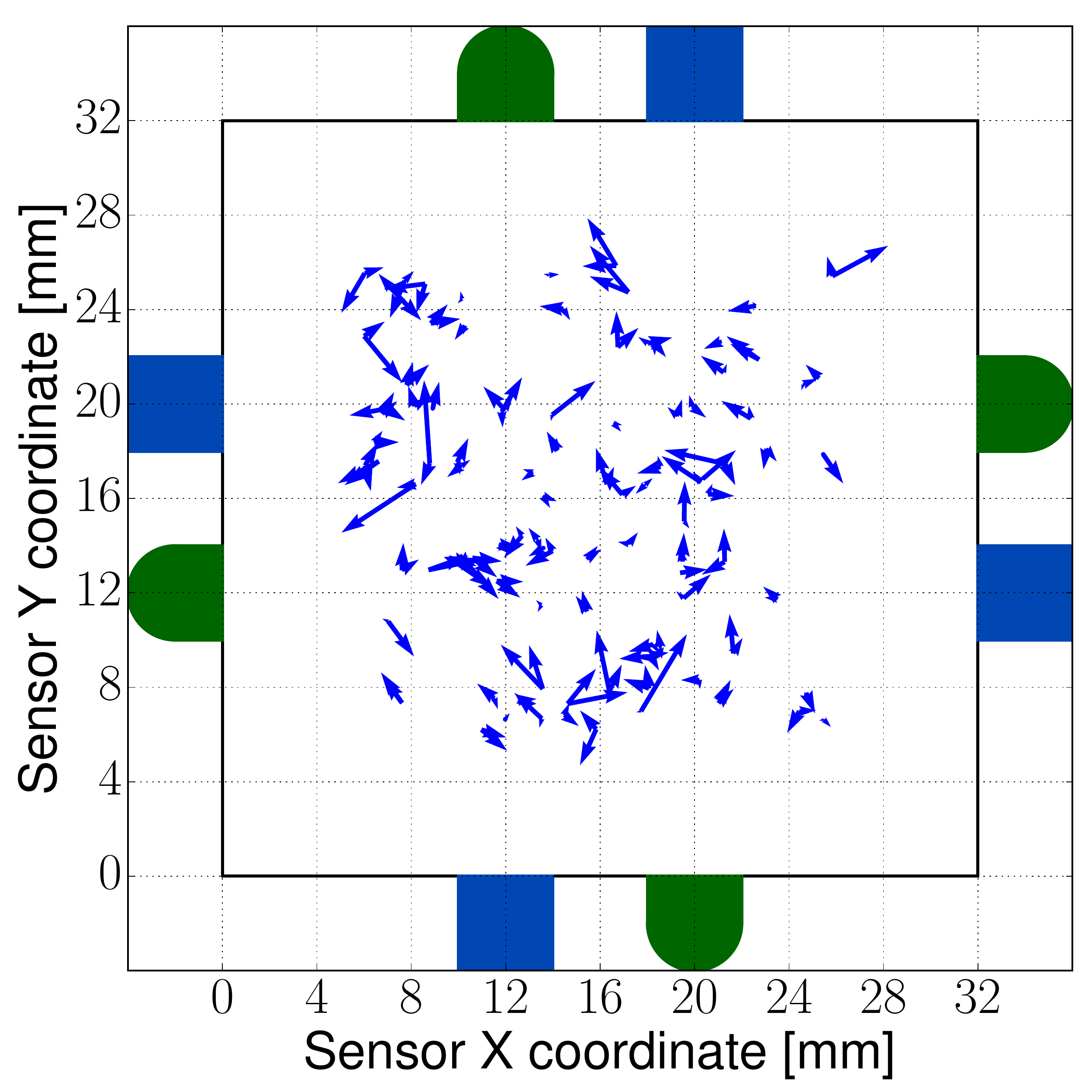}&
\includegraphics[width=0.5\columnwidth]{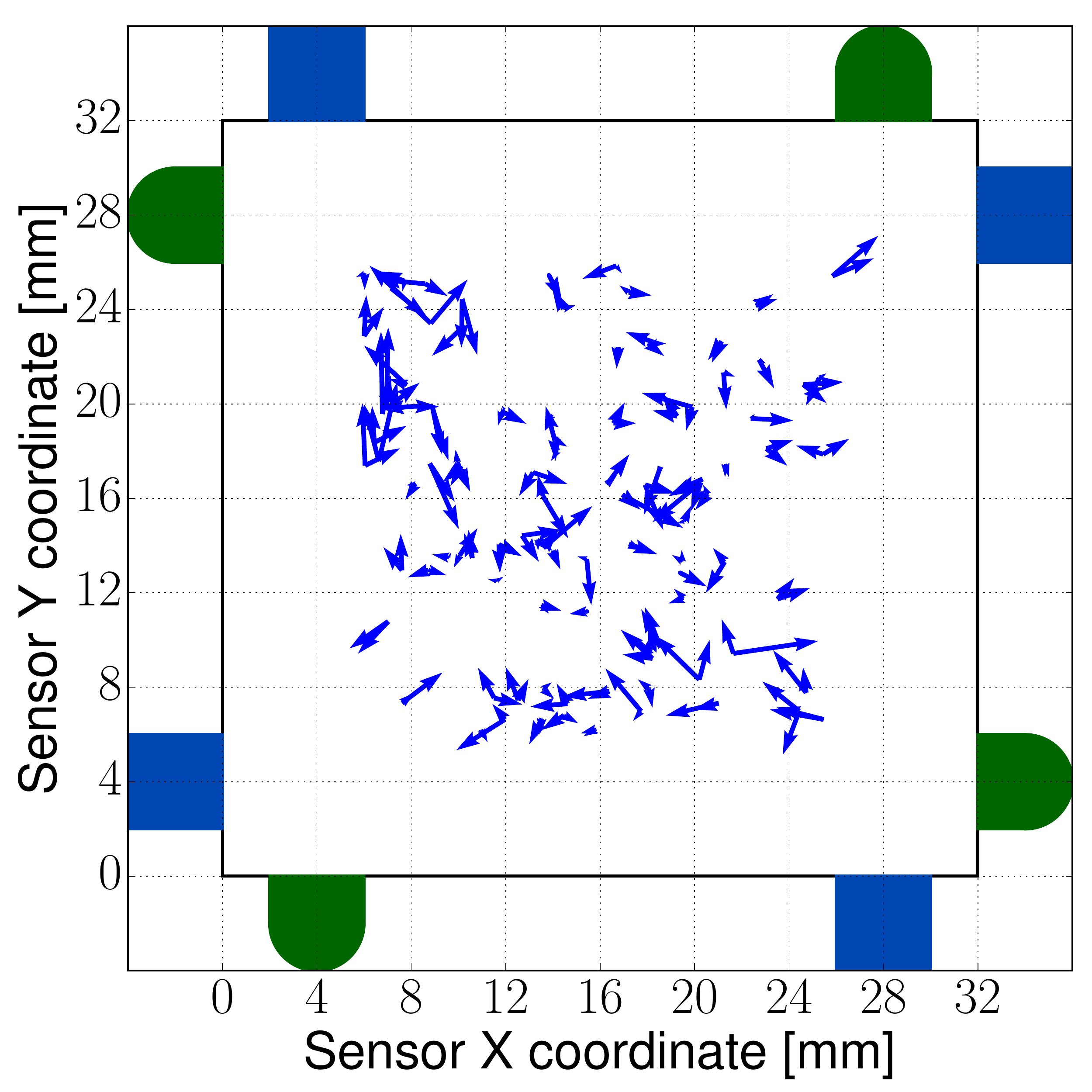}\\[-2mm]
\footnotesize (c) Case 3&
\footnotesize (d) Case 4
\end{tabular}
\caption{Localization error at 2mm indentation depth for cases where
  sensing terminals have been removed from our base
  configuration. Green semicircles (diodes) and blue squares (LEDs)
  indicate the terminals present in each case.}
\label{fig:Sig_Removal}
\end{figure}

\begin{table}[t]
\small
\centering
\caption{Performance based on number and distribution of sensing
  terminals (localization at 2mm depth). Baseline (first row) consists
  of using all available terminals.}
\label{sig_removal_table}
\setlength{\tabcolsep}{1.5mm}
\begin{tabular}{lcccccc}
\hline
\\[-3mm]
\multicolumn{1}{l|}{\multirow{2}{*}{\textbf{Cases}}} & \multicolumn{3}{c|}{\textbf{Localization Error} (mm)}                   & \multicolumn{3}{c}{\textbf{Depth Error} (mm)}      \\
\multicolumn{1}{l|}{}                                & \textbf{Median} & \textbf{Mean} & \multicolumn{1}{c|}{\textbf{Std Dev}} & \textbf{Median} & \textbf{Mean} & \textbf{Std Dev} \\
\hline
\\[-3mm] \hline
\\[-2mm]
Basel.                                               & 0.42                & 0.50              & 0.31                                 & 0.04                & 0.04              & 0.03            \\
Case 1                                               & 0.62                & 0.69              & 0.41                                 & 0.06                & 0.08              & 0.07            \\
Case 2                                               & 0.70                & 0.80              & 0.49                                 & 0.11                & 0.13              & 0.11            \\
Case 3                                               & 0.86                & 1.02              & 0.67                                 & 0.12                & 0.15              & 0.11            \\
Case 4                                               & 1.11                & 1.24              & 0.75                                 & 0.17                & 0.21              & 0.17            
\end{tabular}
\end{table}

The four cases are presented in Fig.~\ref{fig:Sig_Removal}. Cases 1
and 2 have four sensing terminals removed (4 LEDs in Case 1, 2 LEDs and 2 photodiodes in Case 2). However, they present different distribution of
terminals along the edge of the sensing area. Cases 3 and 4 each have
eight terminals removed (4 LEDs and 4 photodiodes), again with
different spatial distributions. Localization results are also shown
in Fig.~\ref{fig:Sig_Removal}. (Comparative performance for the base
case is in Fig.~\ref{fig:Light_results} at 2mm depth.) Accuracy in
both localization and depth predictions is also summarized in Table
~\ref{sig_removal_table}.

Results show that a higher number of sensing terminals always results
in a lower median error for localization accuracy and for depth
accuracy. Additionally, we make two observations on this data. First,
performance tends to increase or decrease uniformly, over the entire
surface of the sensor, rather than locally, in the vicinity of
additional terminals. Case 2 shows no noticeable difference in
performance along the top vs. bottom edges; Cases 3 and 4 do not
differentiate between corners and edge centers despite different
terminal placements.

Second, for the same number of terminals, a more homogeneous
distribution uniformly increases general performance. Numerical
results show that case 1 performs better than case 2 in both metrics
with the same number of sensing terminals, but with a more homogeneous
distribution of the receptive fields for the sensing pairs. Following
the same pattern, case 3 also outperforms case 4. Both of these
phenomena could be due to the fact that our intuition with respect to
how the light travels through the elastomer and what the receptive
fields for a given pair look like might be a bit too simplistic. As
that we rely of a purely data driven method, all of these complex
interactions are captured at training time. Future studies can provide
more insight as to how to optimize the distribution of the sensing
terminals to aid the design of our sensors.

\mystep{Multistage training for larger sensors.} An important question
regarding our approach is its ability to scale to larger sensor
sizes. With our current geometry, increasing the dimension of the
sensor side leads to a proportional increase in the distances between
sensing elements (as long as they are still distributed only on the
border), and a quadratically increased sensing surface area.

To investigate these effects, we constructed a second optical sensor
with a 45 mm side. Accounting for the tip diameter and the safety
margin (as before), this implies a 1024 mm$^2$ sensing area, a 2.56X
increase compared to the sensor shown in Fig.~\ref{fig:three_sensors},
Middle. All other aspects of the design where unchanged, including the
number of terminals and their distribution in the sensor.

We applied the exact same data-driven mapping algorithm described in
the previous section. We recall that the core learning algorithm used
is an kernel ridge regressor trained to simultaneously predict indentation
depth and location. We thus refer to this method as ``single-stage
training''; the results on the larger sensor are shown in
Table~\ref{table:multistage}. As expected, we notice a performance
degradation compared to the smaller sensor, even though localization
error is still below 1 mm at indentation depths of 2 mm or higher.

\begin{table}[t]
\small
\centering
\caption{Localization error (mm) for large sensor}
\label{table:multistage}
\hspace{-3mm}
\setlength{\tabcolsep}{1.5mm}
\begin{tabular}{lcccccc}
\hline
\\[-3mm]
\multicolumn{1}{c|}{\textbf{Depth}} & \multicolumn{3}{|c}{\textbf{Single stage training}} & \multicolumn{3}{|c}{\textbf{Multistage training}}      \\
\multicolumn{1}{c|}{(mm)}               & \multicolumn{1}{c}{\textbf{Median}} &  \multicolumn{1}{c}{\textbf{Mean}} & \multicolumn{1}{c|}{\textbf{Std. Dev}}  & \multicolumn{1}{c}{\textbf{Median}} &  \multicolumn{1}{c}{\textbf{Mean}} & \multicolumn{1}{c}{\textbf{Std. Dev}} \\ \hline
\\[-3mm] \hline
\\[-2mm]
0.1  & 2.22  & 2.63  & 1.87  & 1.72  & 2.15  & 1.65  \\
0.5  & 1.67  & 1.95  & 1.29  & 1.27  & 1.48  & 1.03  \\
1.0  & 1.36  & 1.50  & 0.92  & 1.17  & 1.32  & 0.84  \\
2.0  & 0.94  & 1.06  & 0.64  & 0.53  & 0.68  & 0.43  \\
3.0  & 0.74  & 0.81  & 0.47  & 0.63  & 0.75  & 0.49  \\
5.0  & 0.94  & 1.06  & 0.64  & 0.53  & 0.60  & 0.38 
\end{tabular}
\end{table}

One possibility to recover some of the performance drop would be to
increase the amount of training data; however, given the larger
surface size, this becomes prohibitively expensive from a
computational standpoint if training a single regressor. To account
for this, we split our training / testing procedure into specialized
stages. First, we trained an linear regressor to exclusively predict the
depth of the indentation. Second, we used five separate regressors to
predict indentation location, each trained for specific range of
depths. Each of these regressors was trained on a 1 mm slice of
indentation depths, centered at 0.5, 1.5, 2.5, 3.5 and 4.5 mm
respectively.

The corresponding multistage testing procedure is as follows: first,
the specialized depth regressor predicts indentation depth, referred
to as $d_p$. Then, based on this result, the test tuple is sent to the
localization regressor whose range covers $d_p$. In turn, this chosen
regressor predicts the location of the indentation, which completes
our prediction.

The advantage compared to single-stage prediction is that each
localization regressor is only trained on a subset of the data - the
slice corresponding to its assigned depth range. We can thus afford to
increase the resolution of our training samples by a similar amount
with no effect on the training time. Taking advantage of this, we
trained our predictors using training data sampled at every 0.1 mm in
depth (as opposed to 0.5 mm in the previous section). The results of
this multistage approach (with more training data) are also shown in
Table~\ref{table:multistage}.

We notice that adding more training data increases performance at all
depths and approaches the results obtained for the smaller
sensor. Overall, we believe these results show that our approach can
scale to larger sensor sizes, and that training variants that allow
for more data can, up to a point, help mitigate performance loss. Of
course, if the sensor size keeps increasing, we expect to run into
computational limits that could require a completely different learning
approach. However, we believe that the dimensions demonstrate here are
in line with the requirements of our desired application, integration
into robot fingers for human-scale manipulation.

\subsection{Discussion and limitations}

The perimeter-based THT optical sensor combines sub-millimeter
localization accuracy (in many cases improving to half-millimeter)
with indentation depth determination within a tenth of a
millimeter. Still, these results have all been obtained with a single
indenter tip, used for both training and testing, and do not speak to
the abilty to handle different indenter shapes.

\begin{figure}[t]
\setlength{\tabcolsep}{-1mm}
\begin{tabular}{cc}
\includegraphics[width=0.53\columnwidth]{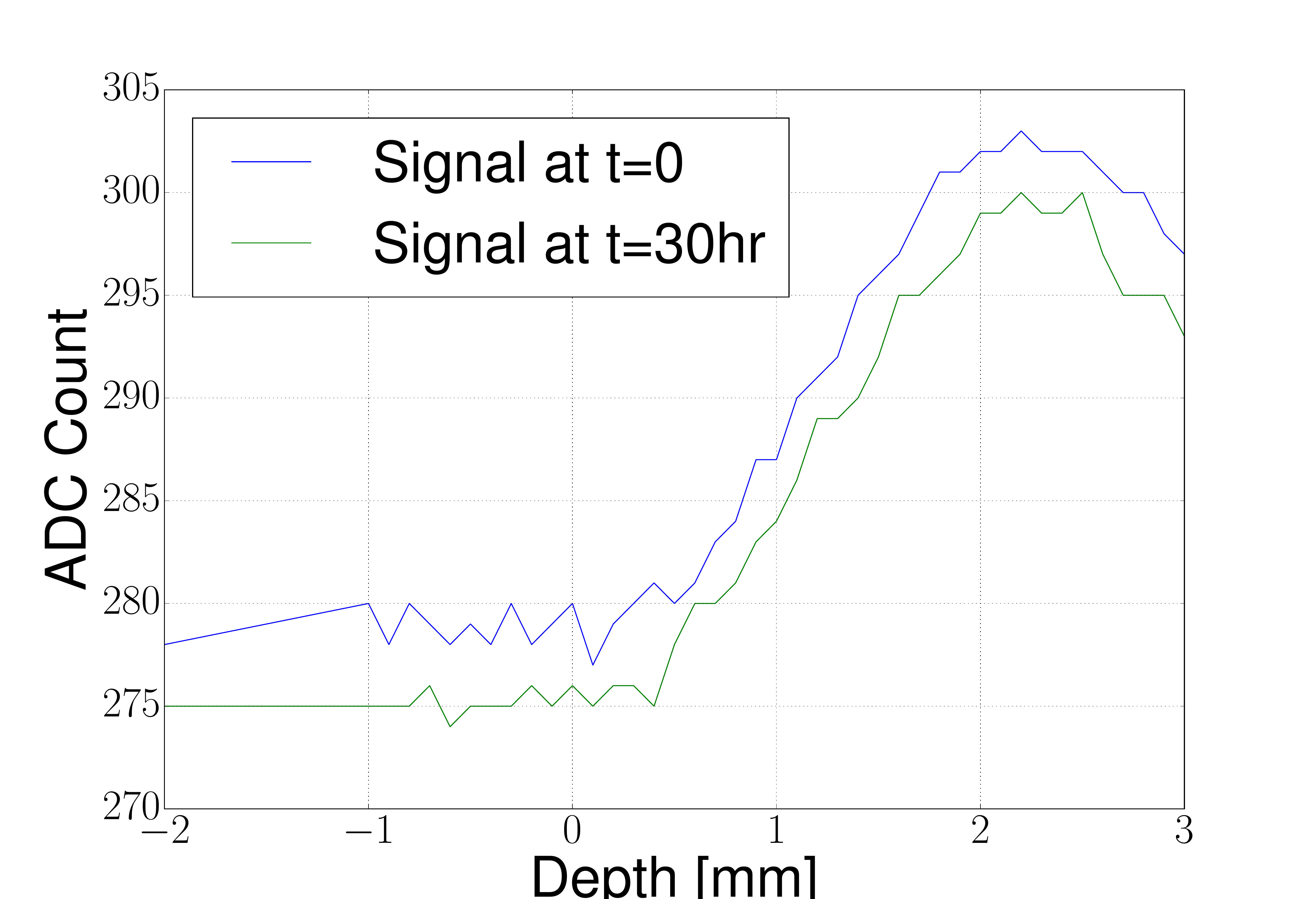}&
\includegraphics[width=0.53\columnwidth]{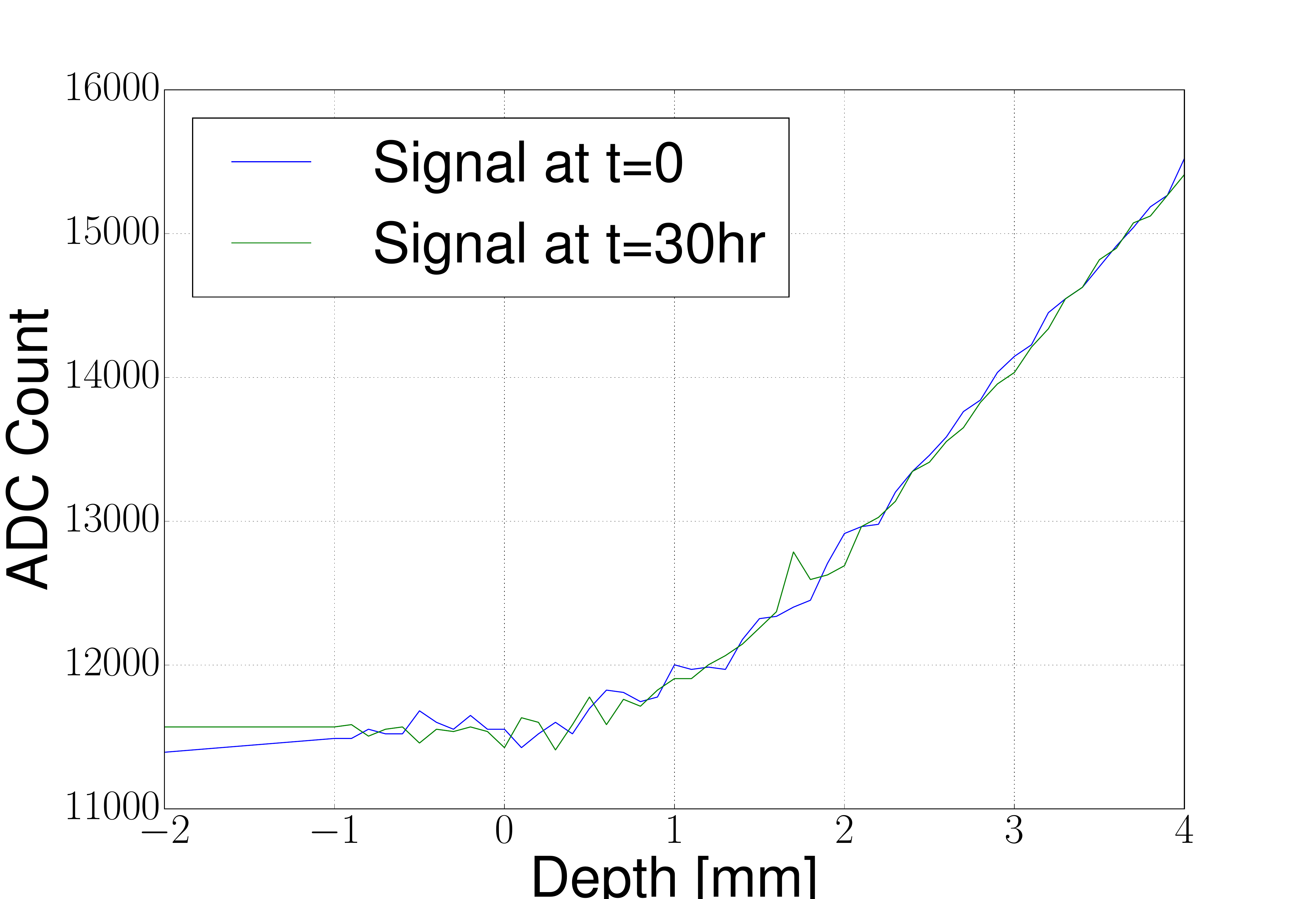}\\[1mm]
\end{tabular}
\caption{Signal comparison over time for two versions of our
  optics-based sensors, with THT components (Left) and SMT
  components (Right). In each case, we performed 60 consecutive
  indentations with the 6mm diameter hemispherical tip in a single sensor location, with the indentations
  spaced 30 minutes apart. Each plot shows the signal from the first
  and last indentation in the respective set.}
\label{fig:drift_analysis}
\end{figure} 

Additionally, we have found that these sensors, while showing no
discernible hysteresis over time scale on the order of seconds, do
exhibit signal drift over periods of hours and large numbers of
repeated indentations (Fig.~\ref{fig:drift_analysis}, Left). Visual
inspection showed that, after repeated indentations and over long
periods of time, the transparent elastomer begins detaching from the
surface of the LEDs, which affects light propagation. We address both
of these aspects with a second version of our sensor, described in the
next section.

\section{Surface-mount optical sensor with base terminals}

For the second iteration of our optics-based sensor, we switched from
through-hole technology (THT) LEDs and photodiodes to surface-mount technology (SMT)
components. We developed prototype boards containing two LEDs and two
photodiodes each, with all wiring for the four components combined
through a single connector. Compared to using through-hole
components, this packaging simplified the manufacturing process for
also including terminals on the base of the sensor. This prototype is
shown in Fig.~\ref{fig:three_sensors}, Right.

\subsection{Sensor manufacture}

The SMT optic sensor follows a similar fabrication method to the THT version.
We initially 3D print out of ABS plastic a square mold with exterior dimensions of 48mm x 48mm. 
The interior cavity in the mold is 38mm x 38mm. This sensor presents sockets both on its walls, as well as 
on its base to fit the small PCB boards where the LEDs and photodiodes are located. 

Each PCB board measures 29mm length and 8mm width. The front of the PCB board holds two 1.6 mm x 0.8 mm LEDs (Kingbright APG1608ZGC, peak wavelength 515nm)
and two 4.0mm x 4.5mm photodiodes (OSRAM BPW34S E9601, peak sensitivity 850nm) in alternate pattern. Additionally, the front holds an FFC connector 
to interface with our measuring circuit. The back of the PCB board holds an operational amplifier 
(Analog Devices AD8616) on a trans-impedance amplifier configuration. Overall this sensor uses 7 PCB boards, 4 being located on 
the mold's walls and the remaining 3 being placed on the mold's base. This represents a total of 14 LEDs and 14 photodiodes.

For the same reasons as in our THT optic sensor, before installing the PCB boards, we coat the base of the sensor with a 1mm
layer of PDMS saturated with carbon black particles. After curing this layer, we cut out the carbon black infused PDMS on the sockets so that 
we can place the boards inside them. For all PCB boards we use a few drops of glue to place them inside their sockets. 
The flexible flat cables used to connect to each individual board are routed through small holes on the walls of the 3D printed mold. 
After connecting each board, these holes are sealed using epoxy. At this point, we fill the mold cavity with the elastomer at a 1:20 curing agent to PDMS ratio. 

A microcontroller (NXP LPC1768) handles switching the LEDs and taking analogue readings of each photodiode. The resulting sampling rate to collect and report 
all signals is approximately 100Hz. 

An important result of this manufacturing process is the absence of
drift over large time scales and large numbers of intermediate
indentations. Fig.~\ref{fig:drift_analysis} shows a comparison of
individual signals over time for both versions of the optical
sensor. The THT sensor exhibits a clear drift in the
signal while the SMT sensor shows no similar phenomenon. We attribute
this change to the use of SMT LEDs with much smaller surface area,
that do not exhibit unwanted bonding/unbonding behavior with the
elastomer.

\begin{figure}[t]
\setlength{\tabcolsep}{0mm}
\begin{tabular}{cc}
\includegraphics[clip, trim={35cm 30cm 35cm 30cm}, width=0.5\columnwidth]{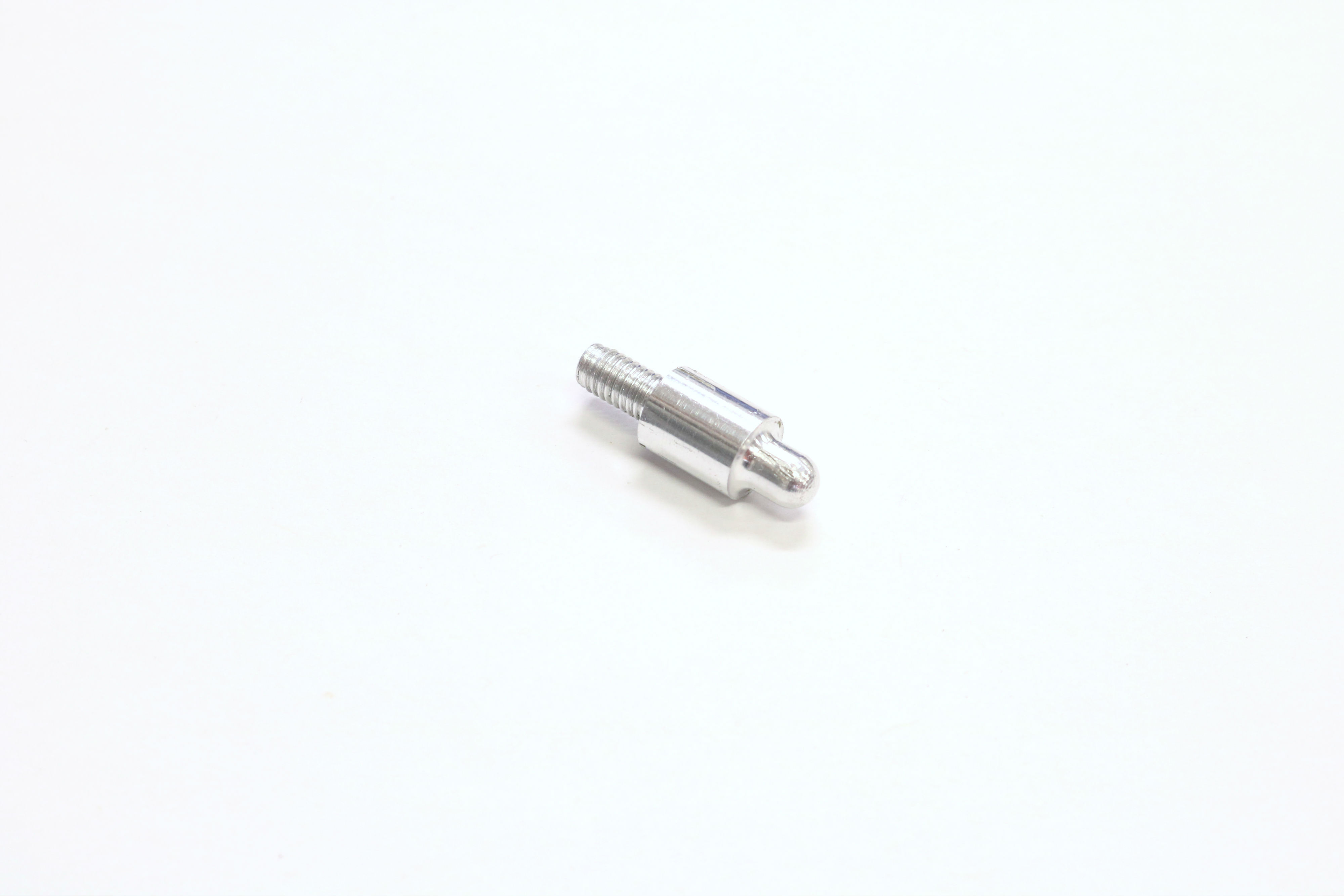}&
\includegraphics[clip, trim={35cm 30cm 35cm 30cm}, width=0.5\columnwidth]{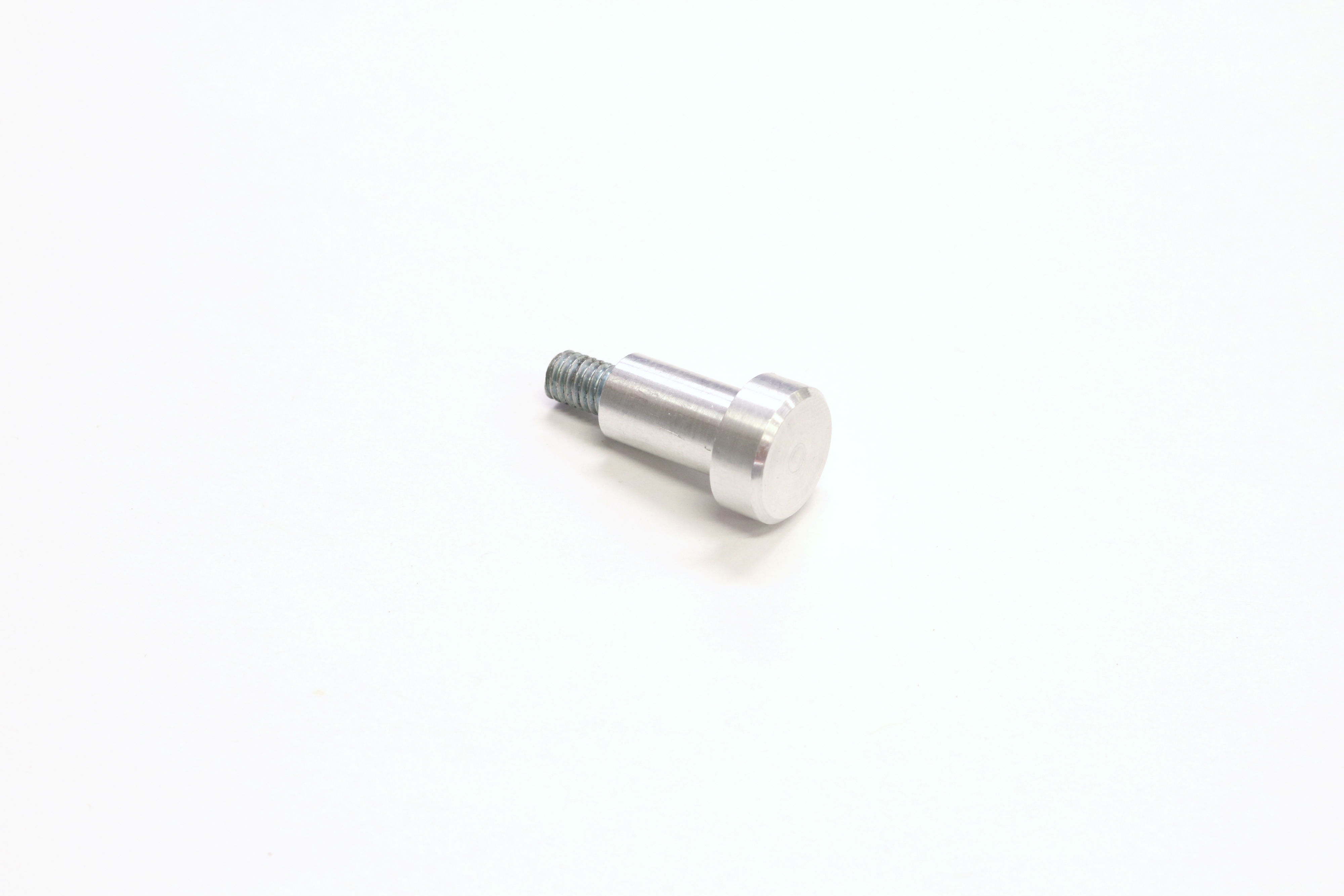}\\[-2mm]
% \footnotesize (a) Spherical Tip&
% \footnotesize (b) Flat Tip\\[2mm]
\includegraphics[clip, trim={35cm 30cm 35cm 30cm}, width=0.5\columnwidth]{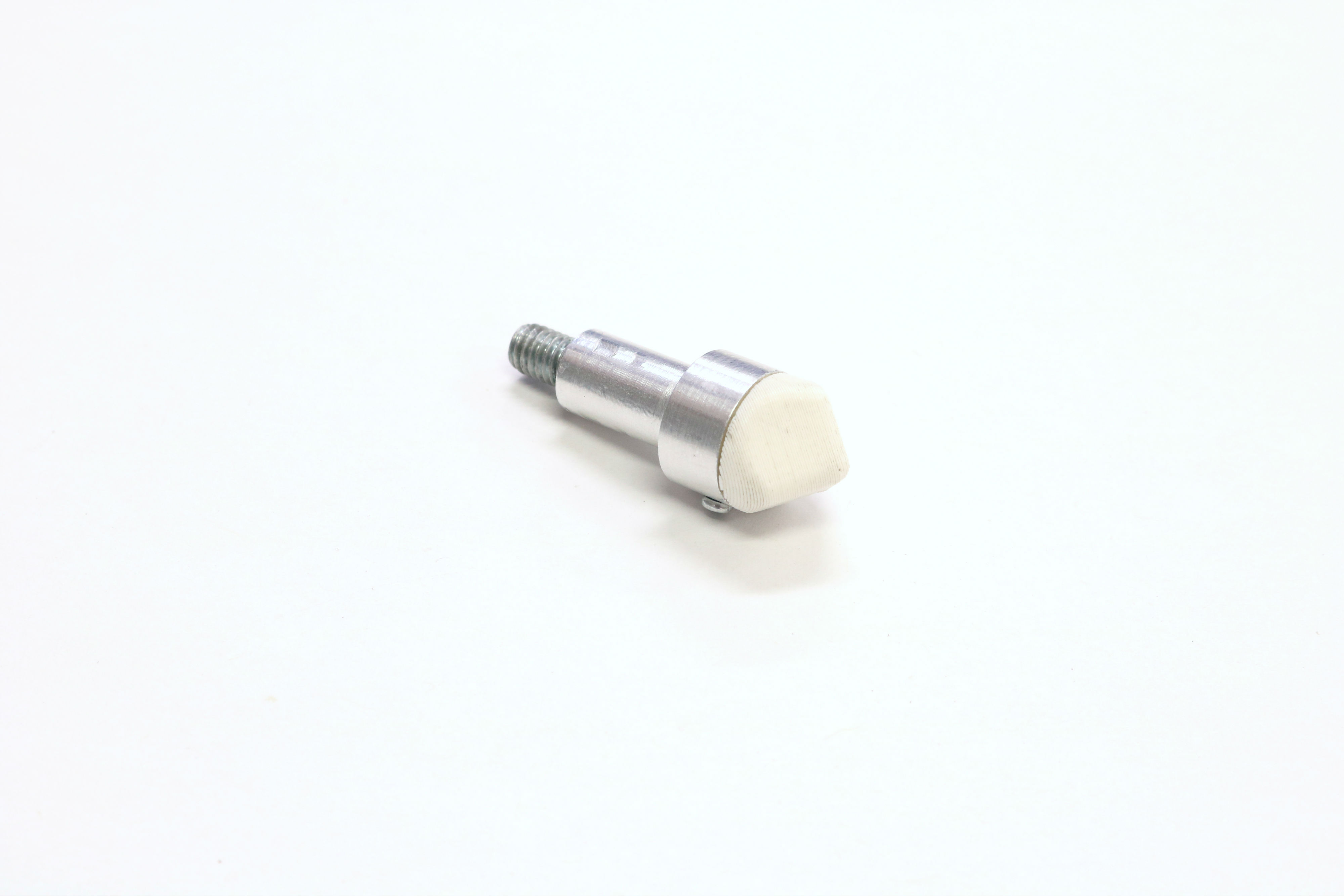}&
\includegraphics[clip, trim={35cm 30cm 35cm 30cm}, width=0.5\columnwidth]{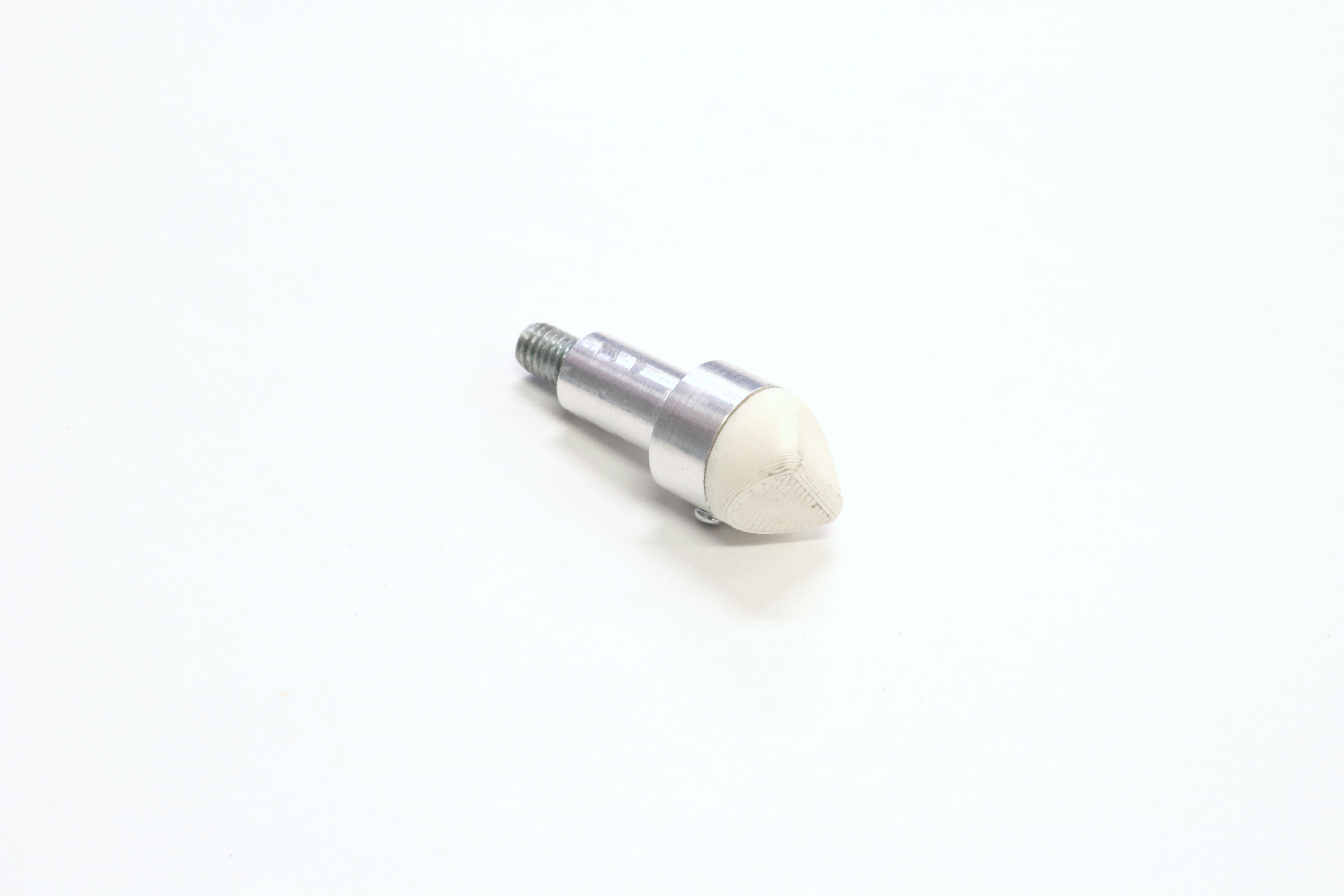}\\[-2mm]
% \footnotesize (c) Corner Tip 3&
% \footnotesize (d) Edge Tip
\end{tabular}
\caption{The four different indenter types used for our classification: hemispherical, flat, corner and edge tips.
The edge tip (lower right) was used with orientations of 0, 120 and 240 degrees.}
\label{fig:Indenter_Geometries}
\end{figure}

\subsection{Data collection}
We collect data on this sensor in the same manner as that presented
for the first iteration of our optics-based sensor presented in 
Section~\ref{sec:optics_collection}. One difference is that we now
use 14 LEDs and 14 photodiodes. We continue to read a baseline 
signal with all LEDs turned off which we subtract from all other 
signals to make the sensor robust under changing ambient light 
conditions. This results in 15 signals for each photodiode.

The primary difference with this sensor is that we now repeat this data
collection process for multiple indenter tips. To mimic geometries 
which are likely to be encountered in common workspaces, we manufactured
the following indenter tips: circular planar with $15mm$ diameter, 
$90^{\circ}$ edge with $15mm$ length, and a square corner with maximum
width $15mm$. We take separate data sets with the edge tip oriented at
three positions, each $120^{\circ}$ apart. We also include the $6mm$
diameter hemispherical tip which was used in previous sections. 
This results in 6 different indenter tips.

Following the depth convention outlined in Section~\ref{sec:optics_collection}, 
we take one measurement at $-10mm$ and measurements at $1mm$ intervals between 
depths of $-5mm$ and $-1mm$. We then take a measurement every $0.1mm$ down to 
the maximum indendation depth. The process is mirrored on retraction. Due to 
force constraints on our indentation actuator, every tip does not reach the 
same maximum indentation depth. The planar tip reaches a maximum depth of 
$1.2mm$, the edge tip reaches a maximum of $3.0mm$, and all other tips reach 
a maximum depth of $4.0mm$.

Each measurement $i$ now results in a tuple of the form
$\Phi_i=(t_i,x_i,y_i,d_i,p_{j=1}^1,..,p_{j=1}^{14},...,p_{j=15}^1,..,p_{j=15}^{14})$
where $t_i \in \{1,..,6\}$ identifies the tip used for the
indentation, $(x_i,y_i)$ is the indentation location in sensor
coordinates, $d_i$ is the depth at which the measurement was taken and
$(p_{j}^1,..,p_{j}^{14})$ correspond to readings of our 14 photodiodes
as we turn each LED on: state $j\ \epsilon\ [1,14]$, from which we
subtract the ambient light captured by each diode when all LEDs are
turned off (state $j=15$). We thus have a total of 214 numbers in each
tuple $\Phi_i$.

We again use a \textit{grid indentation pattern} for training and a 
\textit{random indentation pattern} for testing. This sensor has a 
$38mm$ x $38mm$ PDMS surface. Taking into account our larger tip 
radius ($7.5mm$) and a $3.5mm$ margin, our indentation pattern results 
in 81 locations distributed in $2mm$ intervals over a $16mm$ x $16mm$ 
grid. For testing, the \textit{random indentation pattern} consists 
of 100 indentation events scattered within the same area.

\subsection{Analysis and results}
Our objective here is twofold: to determine whether our signals 
contain sufficient information to reliably classify the tip geometry 
with which the sensor is being contacted; and to explore our sensor's 
ability to identify contact depth and location regardless of tip 
geometry. To train our models we use a single data set for each 
tip, each collected in the dark. The feature space for training 
has a dimensionality of 196.  

\begin{figure}[t]
\centering 
\includegraphics[clip, trim=0.4cm 0.4cm 0.4cm 0.4cm, width=0.85\linewidth] {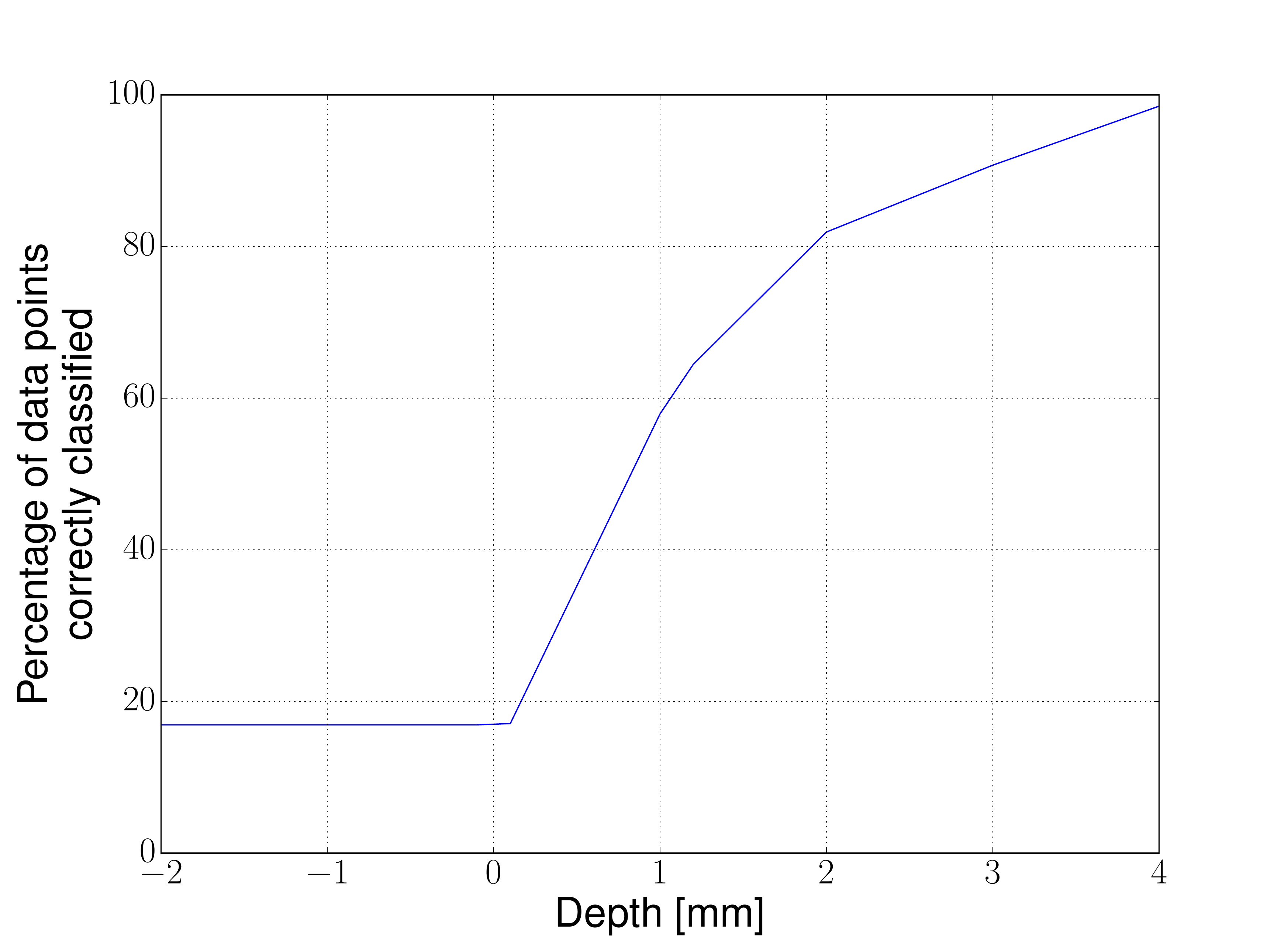}
\caption{Tip classification results over all locations in our test
  dataset as a function of indentation depth. Values in the graph represent the percentage of data points in our test dataset whose tip geometries were correctly classified.}
\label{fig:73_tipClass}
\end{figure}

\mystep{Discriminating between indenter geometries.} To perform this 
classification we use a Neural Network with one input layer, one hidden 
layer (1024 nodes, ReLU activation function), and one output layer 
(6 nodes, softmax activation function). The classifier is trained and 
tested on all six data sets, with both $d_i<0$ and $d_i\geq0$. The 
primary metric for evaluating classifier performance is accuracy; 
that is, the ratio of the number of data points correctly classified 
to the total number of data points. This performance is shown as a 
function of indentation depth in Fig.~\ref{fig:73_tipClass}.

In the limit of negative depths, the classification accuracy converges 
to the expected $16.6\%$, or $1/6$ random chance. The accuracy 
increases with indentation depth and achieves a maximum of $98.5\%$ at 
$4.0mm$. 

\mystep{Depth and location regression.} The goal here is to determine 
our ability to identify contact location and indentation depth regardless
of indenter tip geometry. To do this, we use a kernelized ridge 
regressor analogous to that presented in Section~\ref{sec:optics_results}. 
We train this regressor on all six datasets. Due to the size of this 
combined dataset and the computational requirements of training this 
regressor, we undersample our data based upon depth, maintaining 
higher granularity near the sensor surface. Results presented in 
this section were obtained with $\gamma = 10.0e^{-7}$ and $\alpha = 0.01$. 
The results of testing this regressor on all six datasets are tabulated 
in Table~\ref{table_73_allTips}. Compared to the localization and depth 
regression performance of the optics-based sensor shown in 
Table~\ref{light_table}, we see improved performance at low depths 
and weaker performance at larger depths. However, we consider these 
results to be well within the operating margin for a practicable sensor.

\begin{table}[t]
\small
\centering
\caption{Localization and depth accuracy for all indenter geometries.}
\label{table_73_allTips}
\setlength{\tabcolsep}{1.5mm}
\begin{tabular}{ccccccc}
\hline
\\[-3mm]
\multicolumn{1}{c|}{\textbf{Depth}} & \multicolumn{3}{|c}{\textbf{Localization Error} (mm)}                      & \multicolumn{3}{|c}{\textbf{Depth Error} (mm)}                             \\
\multicolumn{1}{c|}{(mm)}               & \multicolumn{1}{c}{\textbf{Median}} &  \multicolumn{1}{c}{\textbf{Mean}} & \multicolumn{1}{c|}{\textbf{Std. Dev}}  & \multicolumn{1}{c}{\textbf{Median}} &  \multicolumn{1}{c}{\textbf{Mean}} & \multicolumn{1}{c}{\textbf{Std. Dev}} \\ \hline
\\[-3mm] \hline
\\[-2mm]
0.1          & 1.58                & 1.81              & 1.17              & 0.17                & 0.18              & 0.10              \\
0.5          & 1.35                & 1.53              & 1.00              & 0.10                & 0.12              & 0.09              \\
1.0          & 1.13                & 1.38              & 0.95              & 0.10                & 0.13              & 0.10              \\
2.0          & 0.96                & 1.21              & 0.90              & 0.18                & 0.22              & 0.17              \\
3.0          & 0.91                & 1.14              & 0.85              & 0.28                & 0.33              & 0.24              \\
4.0          & 1.11                & 1.27              & 0.78              & 0.63                & 0.66              & 0.32             
\end{tabular}
\end{table}

\mystep{Tip removal analysis.} In practice, a tactile sensor will
encounter indentation geometries which have not been included in the
training of its working model. In pursuit of robustness when faced
with such a situation, we analyze our sensor's performance when
testing on tip data not included in training. We train six regressors,
each on a different five of the six possible tips.  Each of these
regressors is then tested exclusively on data associated with the
missing tip. The results of this analysis are tabulated in
Table~\ref{table_73_tipRemoval}. Overall, these results are on the
same order as those seen when all tips are included
(Table~\ref{table_73_allTips}), suggesting the ability of our sensor
to accurately predict on indentation geometries not formerly
encountered.

\begin{table}[t]
\small
\centering
\caption{Localization and depth accuracy with individual tips removed
  from training. All results are shown for an indentation depth of
  $2mm$, with the exception of the Planar tip for which results are
  shown at $1mm$ depth.}
\label{table_73_tipRemoval}
%\begin{tabular}{cc}
%\hspace{-6mm}
\setlength{\tabcolsep}{0.9mm}
\begin{tabular}{ccccccc}
\hline
\\[-3mm]
\multicolumn{1}{c|}{\textbf{Tip}} & \multicolumn{3}{|c}{\textbf{Localization Error} (mm)}                      & \multicolumn{3}{|c}{\textbf{Depth Error} (mm)}                             \\
\multicolumn{1}{c|}{\textbf{Removed}}               & \multicolumn{1}{c}{\textbf{Median}} &  \multicolumn{1}{c}{\textbf{Mean}} & \multicolumn{1}{c|}{\textbf{Std. Dev}}  & \multicolumn{1}{c}{\textbf{Median}} &  \multicolumn{1}{c}{\textbf{Mean}} & \multicolumn{1}{c}{\textbf{Std. Dev}} \\ \hline
\\[-3mm] \hline
\\[-2mm]
Planar          & 1.30                & 1.31              & 0.61              & 0.18                & 0.21              & 0.17              \\
Edge 1          & 1.27                & 1.32              & 0.56              & 0.61                & 0.58              & 0.24              \\
Edge 2          & 0.97                & 1.10              & 0.66              & 0.29                & 0.29              & 0.19              \\
Edge 3          & 0.61                & 0.80              & 0.59              & 0.16                & 0.18              & 0.13              \\
Corner          & 0.90                & 1.05              & 0.66              & 1.08                & 1.08              & 0.10              \\
Hemisphere          & 1.10                & 1.36              & 0.91              & 0.51                & 0.51              & 0.11             
%\end{tabular}
\end{tabular}
\end{table}

\subsection{Discussion}

The SMT-based optical sensor adds the important capability to
discriminate between different indenter shapes, while maintaining
accurate localization and depth determination capabilities. An
important characteristic is the lack of discernible hysteresis or
drift over time scales ranging from seconds to days, and over large
number of indentations. This allows us to collect large data sets with
many indenter tips, needed for the tip discrimination.

Along with the new tip discrimination ability, the SMT version also
exhibits a slight decrease in localization and depth prediction
ability compared to the THT version. We attribute this
to the same decreased size of the light emitting terminal: a smaller
LED means fewer rays hitting any given receiver, in both interaction
modes described in Section~\ref{sec:modes}. In the limit, if both the
emitter and the receiver are reduced to points, both modes become
binary, allowing for just an on/off signal, thus reducing the ability
to discriminate. A possible compromise could use multiple SMT LEDs,
activated together as a unit and in close proximity to each other, to
maintain elastomer bonding, but emulate the larger surface area of the
THT version.

\section{Discussion and Conclusions}

In this paper, we explored a data-driven methodology based on multiple 
pairs of sensing terminals distributed inside a volume of soft material 
to create tactile sensors. By using an all-pairs approach, where a sensing 
terminal is paired with every other possible terminal, we obtain a very 
rich data set using fewer wires. We then mine this data to learn the 
mapping between our signals and the variables characterizing a contact 
event. This spatially overlapping signals methodology can be generalized 
to different transduction methods: here we show sensors based on resistivity 
and optics using both THT and SMT components.

Our resistive sensor with an effective sensing area of $160 mm^2$
discriminates contact location with sub-millimeter median accuracy for
an indentation depth of $3mm$. Our THT optical sensors achieve
sub-millimeter localization accuracy over $400 mm^2$ and $1024 mm^2$
workspaces, for indentation depths ranging between $1mm$ and $5mm$. 
The SMT optic sensor can localize contacts with an accuracy of approximately 
$1mm$ regardless of the indenter geometry over a $574 mm^2$ workspace. 
To achieve similar accuracy, the traditional sensor matrix approach 
would require numerous individual taxels with complex manufacturing, wiring and addressing 
techniques. In contrast, we use four terminals (with one wire each) 
for the resistive sensor, and as few as 16 and 28 terminals (with two 
wires each) for the THT and SMT optic sensors respectively.

For all of our THT optic sensors, we also show identification of indentation depth
accurate to within $0.1mm$ for a large part of its operating range. We
use depth here as a proxy for indentation force, based on a known
stiffness curve for our sensor; it is also possible to adjust the PDMS
layer thickness and stiffness to affect the sensor sensitivity and dynamic range. A stiffer PDMS layer will require a
larger force to activate the second detection mode. 

Comparing the two transduction methods presented here, the number of wires required
by each sensor is significantly different. A single sensing terminal
in the resistive sensor takes one wire, versus two wires needed for an
LED or photodiode in the optical sensor. Moreover, the nature of
measuring resistance means we can pair a sensing terminal with any
other, whereas in the optical sensor case, a pair must be formed by an
LED and a photodiode. This reduces the amount of possible pairs for a
fixed amount of sensing terminals. If optic components are deployed on a board 
(rigid or flexible) then common buses can be used for power and ground, greatly reducing the
wire count for the sensor. 

In general, we have found optics to be a more robust transduction
method. The sensitivity and accuracy of the optic sensors are superior
to those of the resistive sensor. The signals extracted from the
resistive sensor present a lower signal to noise ratio (SNR) and are
inherently less sensitive when subjected to small strains compared to
the optic counterpart. A small deformation on the optical sensor can
result in a big signal change product of the frustrated total internal
reflection effect, as well as the change in surface
normals. Additionally, the resistive sensor, as reported in the
literature~\cite{kappassov2015} and confirmed by our experimental
data, is affected by hysteresis and must rely on baseline
measurements. Still, if different manufacturing methods can mitigate
such phenomena, the piezoresistive transduction method still holds
promise thanks to its intrinsically low wire count.

As was shown in this paper, the distribution of the sensing terminals
plays a fundamental role in the sensor performance. While at this
point the distribution for our sensors was decided based on intuition
and simple heuristics in terms of covering the sensing area
homogeneously, further studies should be conducted based on the used
transduction method to aid the sensor design and optimize the number
of sensing terminals used for a given accuracy requirement.

We have also shown on our SMT optic sensor that our data-driven
approach can be robust to different indenter geometries. The sensor
can distinguish between different indenter types and, maybe even more
importantly, provide accurate localization and depth predictions
even when using a previously unseen indenter.

One of the key aspects we have yet to address is the adaptation of
this method to arbitrary, non-planar geometries encountered in
robot hands; this is one of the main goals of our future work. Previous work
by To et al.~\cite{to2015} shows it is feasible to guide light through curved 
geometries of an optically clear elastomer with an external reflective layer. 
For this reason, we believe our method can be successfully deployed on a robot hand.
We also plan to investigate how to implement the ability to detect multiple 
simultaneous touch points and the possibility of learning additional variables, such as shear forces or
torsional friction. Sensitivity to environmental factors is also
an important metric that should be further analyzed.

We believe that ultimately the number of variables that can be learned
and how accurately we can determine those variables depends on the raw
data that can be harvested from the sensor. Increasing the number of
sensing units or even incorporating different sensors embedded into
the elastomer can extend the sensing modalities in our
sensor. Different learning methods that also account for temporal
consistency between consecutive readings might also improve the
performance or capability of the sensor. All these ideas will be
explored in future work.

\bibliographystyle{IEEEtran}
\bibliography{bib/tactile_sensing}{}

\end{document}